\documentclass[lettersize,journal]{IEEEtran}
\usepackage{amsmath,amssymb,amsfonts}
\usepackage{algorithmic}
\usepackage{algorithm}
\usepackage{array}
\usepackage[caption=false,font=normalsize,labelfont=sf,textfont=sf]{subfig}
\usepackage{textcomp}
\usepackage{stfloats}
\usepackage{url}
\usepackage{verbatim}
\usepackage{graphicx}
\usepackage{mathrsfs}
\usepackage{cite}
\usepackage{pifont}
\usepackage{hyperref}
\usepackage{multirow}
\usepackage{balance}
\usepackage[table,xcdraw]{xcolor}
\usepackage{mathtools}
\usepackage{orcidlink}
\def\rot{\rotatebox} 
\newcommand{\cmark}{\ding{51}} 
\newcommand{\xmark}{\ding{55}} 
\newcommand{\smark}{\ding{72}} 
\begin{document}

\title{Nigel - Mechatronic Design and Robust Sim2Real Control of an Over-Actuated Autonomous Vehicle}

\author{Chinmay V. Samak$^{\ast}$ \orcidlink{0000-0002-6455-6716}, Tanmay V. Samak$^{\ast}$ \orcidlink{0000-0002-9717-0764}, Javad M. Velni \orcidlink{0000-0001-8546-221X} and Venkat N. Krovi\orcidlink{0000-0003-2539-896X}

\thanks{$^{\ast}$These authors contributed equally.}

\thanks{C. V. Samak, T. V. Samak and V. N. Krovi are with the Automation, Robotics and Mechatronics Laboratory (ARMLab), Department of Automotive Engineering, Clemson University International Center for Automotive Research (CU-ICAR), Greenville, SC 29607, USA. Email: {\tt\small {\{\href{mailto:csamak@clemson.edu}{csamak}, \href{mailto:tsamak@clemson.edu}{tsamak}, \href{mailto:vkrovi@clemson.edu}{vkrovi}\}@clemson.edu}}}

\thanks{J. M. Velni is with the Department of Mechanical Engineering, Clemson University, Clemson, SC 29634, USA. Email: {\tt\small {\href{mailto:javadm@clemson.edu}{javadm@clemson.edu}}}}
}



\maketitle


\begin{abstract}
Simulation to reality (sim2real) transfer from a dynamics and controls perspective usually involves re-tuning or adapting the designed algorithms to suit real-world operating conditions, which often violates the performance guarantees established originally. This work presents a generalizable framework for achieving reliable sim2real transfer of autonomy-oriented control systems using multi-model multi-objective robust optimal control synthesis, which lends well to uncertainty handling and disturbance rejection with theoretical guarantees. Particularly, this work is centered around a novel actuation-redundant scaled autonomous vehicle called Nigel, with independent all-wheel drive and independent all-wheel steering architecture, whose enhanced configuration space bodes well for robust control applications. To this end, we present the mechatronic design, dynamics modeling, parameter identification, and robust stabilizing as well as tracking control of Nigel using the proposed framework, with exhaustive experimentation and benchmarking in simulation as well as real-world settings.
\end{abstract}



\section{Introduction}
\label{Section: Introduction}

\IEEEPARstart{M}{odern} day automotive systems exploit a combination of mechanical electrical, electronic, networking, and software sub-systems to enhance performance via a hierarchical suite of autonomy-oriented control realizations. While earlier autonomy developers may have enjoyed the freedom of primarily focusing on core software development, the present context demands a paradigm shift towards a synergistic hardware-software co-design approach, harmonizing with the principles of mechatronics engineering \cite{deSilva2004}. In particular, the demand for increased maneuverability, enhanced control configuration space, and improved tolerance against faults motivates the pursuit of unconventional vehicle designs. However, with the advent of novel design architectures, advanced control strategies \cite{CS4AV} are required to fully exploit the added capabilities. Moreover, the devised control systems need to successfully transition the sim2real gap by guaranteeing robust performance against real-world uncertainties and disturbances. This further calls for capable cyber-physical deployment platforms of varying scales. Previous works have tried to address some of these aspects in isolation, as we will review below.

\begin{figure}[t]
	\centering
	\includegraphics[width=\linewidth]{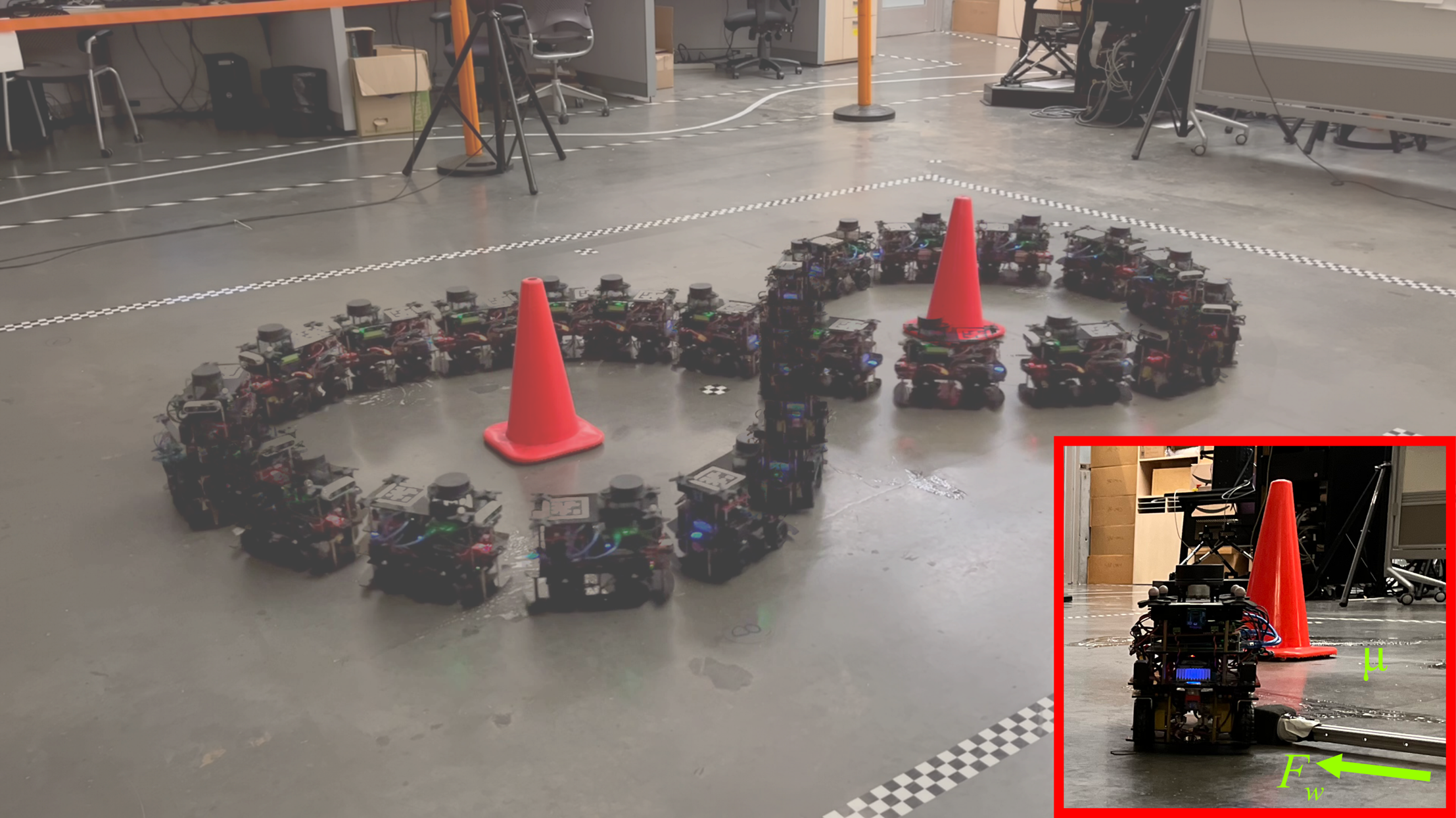}
	\caption{Nigel: a mechatronically redundant and reconfigurable 1:14 scale autonomous vehicle tracking ``figure-8'' maneuver, with the inset depicting disturbance $F_w$ and frictional uncertainties $\mu$ during sim2real transfer.}
	\label{fig1}
\end{figure}

\begin{table*}[t]
\centering
\caption{Comparative analysis of popular scaled autonomous vehicle platforms}
\label{tab1}
\resizebox{\textwidth}{!}{%
    \begin{tabular}{
        >{\columncolor[HTML]{C0C0C0}}l 
        >{\columncolor[HTML]{C6EFCE}}c 
        >{\columncolor[HTML]{FFEB9C}}c 
        >{\columncolor[HTML]{C6EFCE}}c l
        >{\columncolor[HTML]{FFC7CE}}c 
        >{\columncolor[HTML]{FFC7CE}}c c
        >{\columncolor[HTML]{FFC7CE}}c 
        >{\columncolor[HTML]{C6EFCE}}c 
        >{\columncolor[HTML]{C6EFCE}}c 
        >{\columncolor[HTML]{C6EFCE}}l cc
        >{\columncolor[HTML]{C6EFCE}}l 
        >{\columncolor[HTML]{FFC7CE}}l 
        >{\columncolor[HTML]{FFEB9C}}l cc
        >{\columncolor[HTML]{FFC7CE}}c 
        >{\columncolor[HTML]{FFC7CE}}c 
        >{\columncolor[HTML]{FFC7CE}}c 
        >{\columncolor[HTML]{C6EFCE}}c 
        >{\columncolor[HTML]{FFC7CE}}c
        >{\columncolor[HTML]{FFC7CE}}l
        >{\columncolor[HTML]{FFC7CE}}c
        >{\columncolor[HTML]{FFC7CE}}c
        >{\columncolor[HTML]{FFC7CE}}c
        }
        \multicolumn{1}{c}{\multirow{-2}{*}{\cellcolor[HTML]{85DFFF}\textbf{\begin{tabular}[c]{@{}l@{}}\\Platform\end{tabular}}}}
        & \multicolumn{1}{r}{\rot{90}{\cellcolor[HTML]{85DFFF}}}
        & \multicolumn{1}{r}{\rot{90}{\cellcolor[HTML]{85DFFF}}}
        & \multicolumn{1}{r}{\rot{90}{\cellcolor[HTML]{85DFFF}}}
        & \multicolumn{1}{c}{\multirow{-2}{*}{\cellcolor[HTML]{85DFFF}\textbf{\begin{tabular}[c]{@{}l@{}}\\Cost$^{\mathrm{\dagger}}$\end{tabular}}}}
        & \multicolumn{9}{c}{\cellcolor[HTML]{85DFFF}\textbf{Sensing Modalities}}
        & \multicolumn{2}{c}{\cellcolor[HTML]{85DFFF}\textbf{Computational Resources}}
        & \multicolumn{1}{c}{\cellcolor[HTML]{85DFFF}\textbf{\begin{tabular}[c]{@{}c@{}}\\Actuation\\Mechanism$^{\mathrm{\ddagger}}$\end{tabular}}}
        & \multicolumn{2}{c}{\cellcolor[HTML]{85DFFF}\textbf{\begin{tabular}[c]{@{}c@{}}\\Lights \& \\Indicators\end{tabular}}}
        & \multicolumn{2}{c}{\cellcolor[HTML]{85DFFF}\textbf{V2X Support}}
        & \multicolumn{7}{c}{\cellcolor[HTML]{85DFFF}\textbf{API Support}}
        \\
        \multicolumn{1}{c}{\cellcolor[HTML]{85DFFF}}
        & \multicolumn{1}{r}{\rot{90}{\multirow{-2}{*}{\cellcolor[HTML]{85DFFF}\textbf{Scale}}}}
        & \multicolumn{1}{r}{\rot{90}{\multirow{-2}{*}{\cellcolor[HTML]{85DFFF}\textbf{Open Hardware}}}}
        & \multicolumn{1}{r}{\rot{90}{\multirow{-2}{*}{\cellcolor[HTML]{85DFFF}\textbf{Open Software}}}}
        & \multicolumn{1}{c}{\cellcolor[HTML]{85DFFF}}
        & \multicolumn{1}{l}{\rot{90}{\cellcolor[HTML]{85DFFF}\textbf{\textit{Throttle}}}}
        & \multicolumn{1}{l}{\rot{90}{\cellcolor[HTML]{85DFFF}\textbf{\textit{Steering}}}}
        & \multicolumn{1}{l}{\rot{90}{\cellcolor[HTML]{85DFFF}\textbf{\textit{Wheel Encoders}}}}
        & \multicolumn{1}{l}{\rot{90}{\cellcolor[HTML]{85DFFF}\textbf{\textit{GPS/IPS}}}}
        & \multicolumn{1}{l}{\rot{90}{\cellcolor[HTML]{85DFFF}\textbf{\textit{IMU}}}}
        & \multicolumn{1}{l}{\rot{90}{\cellcolor[HTML]{85DFFF}\textbf{\textit{Microphone}}}}
        & \multicolumn{1}{l}{\rot{90}{\cellcolor[HTML]{85DFFF}\textbf{\textit{LIDAR}}}}
        & \multicolumn{1}{l}{\rot{90}{\cellcolor[HTML]{85DFFF}\textbf{\textit{2D Camera}}}}
        & \multicolumn{1}{l}{\rot{90}{\cellcolor[HTML]{85DFFF}\textbf{\textit{3D Camera}}}}
        & \multicolumn{1}{c}{\cellcolor[HTML]{85DFFF}\textbf{\textit{High-Level}}}
        & \multicolumn{1}{c}{\cellcolor[HTML]{85DFFF}\textbf{\textit{Low-Level}}}
        & \multicolumn{1}{c}{\cellcolor[HTML]{85DFFF}}
        & \rot{90}{\cellcolor[HTML]{85DFFF}\textbf{\textit{Lights}}}
        & \rot{90}{\cellcolor[HTML]{85DFFF}\textbf{\textit{Buzzer}}}
        & \cellcolor[HTML]{85DFFF}\textbf{\textit{V2V}}
        & \cellcolor[HTML]{85DFFF}\textbf{\textit{V2I}}
        & \rot{90}{\cellcolor[HTML]{85DFFF}\textbf{\textit{C++}}}
        & \rot{90}{\cellcolor[HTML]{85DFFF}\textbf{\textit{Python}}}
        & \rot{90}{\cellcolor[HTML]{85DFFF}\textbf{\textit{ROS}}}
        & \rot{90}{\cellcolor[HTML]{85DFFF}\textbf{\textit{ROS 2}}}
        & \rot{90}{\cellcolor[HTML]{85DFFF}\textbf{\textit{Autoware}}}
        & \rot{90}{\cellcolor[HTML]{85DFFF}\textbf{\textit{MATLAB}}}
        & \rot{90}{\cellcolor[HTML]{85DFFF}\textbf{\textit{Webapp}}}
        \\
        Nigel (Ours)
        & \cellcolor[HTML]{C6EFCE}{\color[HTML]{006100} 1:14}
        & \cellcolor[HTML]{C6EFCE}{\color[HTML]{006100} \cmark}
        & \cellcolor[HTML]{C6EFCE}{\color[HTML]{006100} \cmark}
        & \cellcolor[HTML]{C6EFCE}{\color[HTML]{006100} \$600}
        & \cellcolor[HTML]{C6EFCE}{\color[HTML]{006100} \cmark}
        & \cellcolor[HTML]{C6EFCE}{\color[HTML]{006100} \cmark}
        & \cellcolor[HTML]{C6EFCE}{\color[HTML]{006100} \cmark}
        & \cellcolor[HTML]{C6EFCE}{\color[HTML]{006100} \cmark}
        & \cellcolor[HTML]{C6EFCE}{\color[HTML]{006100} \cmark}
        & \cellcolor[HTML]{C6EFCE}{\color[HTML]{006100} \cmark}
        & \cellcolor[HTML]{C6EFCE}{\color[HTML]{006100} \cmark}
        & \cellcolor[HTML]{C6EFCE}{\color[HTML]{006100} \cmark}
        & \cellcolor[HTML]{C6EFCE}{\color[HTML]{006100} \cmark}
        & \cellcolor[HTML]{C6EFCE}{\color[HTML]{006100} Jetson Orin Nano}
        & \cellcolor[HTML]{C6EFCE}{\color[HTML]{006100} Arduino Mega}
        & \cellcolor[HTML]{C6EFCE}{\color[HTML]{006100} 4WD4WS}
        & \cellcolor[HTML]{C6EFCE}{\color[HTML]{006100} \cmark}
        & \cellcolor[HTML]{C6EFCE}{\color[HTML]{006100} \cmark}
        & \cellcolor[HTML]{C6EFCE}{\color[HTML]{006100} \cmark}
        & \cellcolor[HTML]{C6EFCE}{\color[HTML]{006100} \cmark}
        & \cellcolor[HTML]{C6EFCE}{\color[HTML]{006100} \cmark}
        & \cellcolor[HTML]{C6EFCE}{\color[HTML]{006100} \cmark}
        & \cellcolor[HTML]{C6EFCE}{\color[HTML]{006100} \cmark}
        & \cellcolor[HTML]{C6EFCE}{\color[HTML]{006100} \cmark}
        & \cellcolor[HTML]{C6EFCE}{\color[HTML]{006100} \cmark}
        & \cellcolor[HTML]{C6EFCE}{\color[HTML]{006100} \cmark}
        & \cellcolor[HTML]{C6EFCE}{\color[HTML]{006100} \cmark}
        \\
        MIT Racecar \cite{MIT-Racecar2017}
        & \cellcolor[HTML]{C6EFCE}{\color[HTML]{006100} 1:10}
        & \cellcolor[HTML]{FFEB9C}{\color[HTML]{9C5700} \smark}
        & \cellcolor[HTML]{C6EFCE}{\color[HTML]{006100} \cmark}
        & \cellcolor[HTML]{FFEB9C}{\color[HTML]{9C5700} \$2,600}
        & \cellcolor[HTML]{FFC7CE}{\color[HTML]{9C0006} \xmark}
        & \cellcolor[HTML]{FFC7CE}{\color[HTML]{9C0006} \xmark}
        & \cellcolor[HTML]{FFC7CE}{\color[HTML]{9C0006} \xmark}
        & \cellcolor[HTML]{FFC7CE}{\color[HTML]{9C0006} \xmark}
        & \cellcolor[HTML]{C6EFCE}{\color[HTML]{006100} \cmark}
        & \cellcolor[HTML]{FFC7CE}{\color[HTML]{9C0006} \xmark}
        & \cellcolor[HTML]{C6EFCE}{\color[HTML]{006100} \cmark}
        & \cellcolor[HTML]{FFEB9C}{\color[HTML]{9C5700} \smark}
        & \cellcolor[HTML]{C6EFCE}{\color[HTML]{006100} \cmark}
        & \cellcolor[HTML]{C6EFCE}{\color[HTML]{006100} Jetson TX2}
        & \cellcolor[HTML]{FFEB9C}{\color[HTML]{9C5700} VESC}
        & \cellcolor[HTML]{FFEB9C}{\color[HTML]{9C5700} AS}
        & \cellcolor[HTML]{FFC7CE}{\color[HTML]{9C0006} \xmark}
        & \cellcolor[HTML]{FFC7CE}{\color[HTML]{9C0006} \xmark}
        & \cellcolor[HTML]{FFEB9C}{\color[HTML]{9C5700} \smark}
        & \cellcolor[HTML]{FFC7CE}{\color[HTML]{9C0006} \xmark}
        & \cellcolor[HTML]{FFC7CE}{\color[HTML]{9C0006} \xmark}
        & \cellcolor[HTML]{FFC7CE}{\color[HTML]{9C0006} \xmark}
        & \cellcolor[HTML]{C6EFCE}{\color[HTML]{006100} \cmark}
        & \cellcolor[HTML]{FFC7CE}{\color[HTML]{9C0006} \xmark}
        & \cellcolor[HTML]{FFC7CE}{\color[HTML]{9C0006} \xmark}
        & \cellcolor[HTML]{FFC7CE}{\color[HTML]{9C0006} \xmark}
        & \cellcolor[HTML]{FFC7CE}{\color[HTML]{9C0006} \xmark}
        \\
        AutoRally \cite{AutoRally2021}
        & \cellcolor[HTML]{FFEB9C}{\color[HTML]{9C5700} 1:5}
        & \cellcolor[HTML]{FFEB9C}{\color[HTML]{9C5700} \smark}
        & \cellcolor[HTML]{C6EFCE}{\color[HTML]{006100} \cmark}
        & \cellcolor[HTML]{FFC7CE}{\color[HTML]{9C0006} \$23,300}
        & \cellcolor[HTML]{FFC7CE}{\color[HTML]{9C0006} \xmark}
        & \cellcolor[HTML]{FFC7CE}{\color[HTML]{9C0006} \xmark}
        & \cellcolor[HTML]{C6EFCE}{\color[HTML]{006100} \cmark}
        & \cellcolor[HTML]{C6EFCE}{\color[HTML]{006100} \cmark}
        & \cellcolor[HTML]{C6EFCE}{\color[HTML]{006100} \cmark}
        & \cellcolor[HTML]{FFC7CE}{\color[HTML]{9C0006} \xmark}
        & \cellcolor[HTML]{C6EFCE}{\color[HTML]{006100} \cmark}
        & \cellcolor[HTML]{FFEB9C}{\color[HTML]{9C5700} \smark}
        & \cellcolor[HTML]{C6EFCE}{\color[HTML]{006100} \cmark}
        & \cellcolor[HTML]{FFEB9C}{\color[HTML]{9C5700} Custom}
        & \cellcolor[HTML]{C6EFCE}{\color[HTML]{006100} Teensy LC}
        & \cellcolor[HTML]{FFEB9C}{\color[HTML]{9C5700} AS}
        & \cellcolor[HTML]{FFC7CE}{\color[HTML]{9C0006} \xmark}
        & \cellcolor[HTML]{FFC7CE}{\color[HTML]{9C0006} \xmark}
        & \cellcolor[HTML]{FFEB9C}{\color[HTML]{9C5700} \smark}
        & \cellcolor[HTML]{FFC7CE}{\color[HTML]{9C0006} \xmark}
        & \cellcolor[HTML]{FFC7CE}{\color[HTML]{9C0006} \xmark}
        & \cellcolor[HTML]{FFC7CE}{\color[HTML]{9C0006} \xmark}
        & \cellcolor[HTML]{C6EFCE}{\color[HTML]{006100} \cmark}
        & \cellcolor[HTML]{FFC7CE}{\color[HTML]{9C0006} \xmark}
        & \cellcolor[HTML]{FFC7CE}{\color[HTML]{9C0006} \xmark}
        & \cellcolor[HTML]{FFC7CE}{\color[HTML]{9C0006} \xmark}
        & \cellcolor[HTML]{FFC7CE}{\color[HTML]{9C0006} \xmark}
        \\
        F1TENTH \cite{F1TENTH2019}
        & \cellcolor[HTML]{C6EFCE}{\color[HTML]{006100} 1:10}
        & \cellcolor[HTML]{FFEB9C}{\color[HTML]{9C5700} \smark}
        & \cellcolor[HTML]{C6EFCE}{\color[HTML]{006100} \cmark}
        & \cellcolor[HTML]{FFEB9C}{\color[HTML]{9C5700} \$3,260}
        & \cellcolor[HTML]{FFC7CE}{\color[HTML]{9C0006} \xmark}
        & \cellcolor[HTML]{FFC7CE}{\color[HTML]{9C0006} \xmark}
        & \cellcolor[HTML]{FFC7CE}{\color[HTML]{9C0006} \xmark}
        & \cellcolor[HTML]{FFC7CE}{\color[HTML]{9C0006} \xmark}
        & \cellcolor[HTML]{FFC7CE}{\color[HTML]{9C0006} \xmark}
        & \cellcolor[HTML]{FFC7CE}{\color[HTML]{9C0006} \xmark}
        & \cellcolor[HTML]{C6EFCE}{\color[HTML]{006100} \cmark}
        & \cellcolor[HTML]{FFEB9C}{\color[HTML]{9C5700} \smark}
        & \cellcolor[HTML]{FFEB9C}{\color[HTML]{9C5700} \smark}
        & \cellcolor[HTML]{C6EFCE}{\color[HTML]{006100} Jetson TX2}
        & \cellcolor[HTML]{FFEB9C}{\color[HTML]{9C5700} VESC 6MkV}
        & \cellcolor[HTML]{FFEB9C}{\color[HTML]{9C5700} AS}
        & \cellcolor[HTML]{FFC7CE}{\color[HTML]{9C0006} \xmark}
        & \cellcolor[HTML]{FFC7CE}{\color[HTML]{9C0006} \xmark}
        & \cellcolor[HTML]{C6EFCE}{\color[HTML]{006100} \cmark}
        & \cellcolor[HTML]{FFC7CE}{\color[HTML]{9C0006} \xmark}
        & \cellcolor[HTML]{FFC7CE}{\color[HTML]{9C0006} \xmark}
        & \cellcolor[HTML]{FFC7CE}{\color[HTML]{9C0006} \xmark}
        & \cellcolor[HTML]{C6EFCE}{\color[HTML]{006100} \cmark}
        & \cellcolor[HTML]{C6EFCE}{\color[HTML]{006100} \cmark}
        & \cellcolor[HTML]{C6EFCE}{\color[HTML]{006100} \cmark}
        & \cellcolor[HTML]{FFC7CE}{\color[HTML]{9C0006} \xmark}
        & \cellcolor[HTML]{FFC7CE}{\color[HTML]{9C0006} \xmark}
        \\
        DSV \cite{DSV2017}
        & \cellcolor[HTML]{C6EFCE}{\color[HTML]{006100} 1:10}
        & \cellcolor[HTML]{FFEB9C}{\color[HTML]{9C5700} \smark}
        & \cellcolor[HTML]{C6EFCE}{\color[HTML]{006100} \cmark}
        & \cellcolor[HTML]{FFEB9C}{\color[HTML]{9C5700} \$1,000}
        & \cellcolor[HTML]{FFC7CE}{\color[HTML]{9C0006} \xmark}
        & \cellcolor[HTML]{FFC7CE}{\color[HTML]{9C0006} \xmark}
        & \cellcolor[HTML]{C6EFCE}{\color[HTML]{006100} \cmark}
        & \cellcolor[HTML]{FFC7CE}{\color[HTML]{9C0006} \xmark}
        & \cellcolor[HTML]{C6EFCE}{\color[HTML]{006100} \cmark}
        & \cellcolor[HTML]{FFC7CE}{\color[HTML]{9C0006} \xmark}
        & \cellcolor[HTML]{C6EFCE}{\color[HTML]{006100} \cmark}
        & \cellcolor[HTML]{C6EFCE}{\color[HTML]{006100} \cmark}
        & \cellcolor[HTML]{FFC7CE}{\color[HTML]{9C0006} \xmark}
        & \cellcolor[HTML]{FFEB9C}{\color[HTML]{9C5700} ODROID-XU4}
        & \cellcolor[HTML]{C6EFCE}{\color[HTML]{006100} Arduino Mega}
        & \cellcolor[HTML]{FFEB9C}{\color[HTML]{9C5700} AS}
        & \cellcolor[HTML]{FFC7CE}{\color[HTML]{9C0006} \xmark}
        & \cellcolor[HTML]{FFC7CE}{\color[HTML]{9C0006} \xmark}
        & \cellcolor[HTML]{FFC7CE}{\color[HTML]{9C0006} \xmark}
        & \cellcolor[HTML]{FFC7CE}{\color[HTML]{9C0006} \xmark}
        & \cellcolor[HTML]{FFC7CE}{\color[HTML]{9C0006} \xmark}
        & \cellcolor[HTML]{FFC7CE}{\color[HTML]{9C0006} \xmark}
        & \cellcolor[HTML]{C6EFCE}{\color[HTML]{006100} \cmark}
        & \cellcolor[HTML]{FFC7CE}{\color[HTML]{9C0006} \xmark}
        & \cellcolor[HTML]{FFC7CE}{\color[HTML]{9C0006} \xmark}
        & \cellcolor[HTML]{FFC7CE}{\color[HTML]{9C0006} \xmark}
        & \cellcolor[HTML]{FFC7CE}{\color[HTML]{9C0006} \xmark}
        \\
        MuSHR \cite{MuSHR2019}
        & \cellcolor[HTML]{C6EFCE}{\color[HTML]{006100} 1:10}
        & \cellcolor[HTML]{FFEB9C}{\color[HTML]{9C5700} \smark}
        & \cellcolor[HTML]{C6EFCE}{\color[HTML]{006100} \cmark}
        & \cellcolor[HTML]{C6EFCE}{\color[HTML]{006100} \$930}
        & \cellcolor[HTML]{FFC7CE}{\color[HTML]{9C0006} \xmark}
        & \cellcolor[HTML]{FFC7CE}{\color[HTML]{9C0006} \xmark}
        & \cellcolor[HTML]{FFC7CE}{\color[HTML]{9C0006} \xmark}
        & \cellcolor[HTML]{FFC7CE}{\color[HTML]{9C0006} \xmark}
        & \cellcolor[HTML]{FFC7CE}{\color[HTML]{9C0006} \xmark}
        & \cellcolor[HTML]{FFC7CE}{\color[HTML]{9C0006} \xmark}
        & \cellcolor[HTML]{C6EFCE}{\color[HTML]{006100} \cmark}
        & \cellcolor[HTML]{FFEB9C}{\color[HTML]{9C5700} \smark}
        & \cellcolor[HTML]{C6EFCE}{\color[HTML]{006100} \cmark}
        & \cellcolor[HTML]{C6EFCE}{\color[HTML]{006100} Jetson Nano}
        & \cellcolor[HTML]{FFEB9C}{\color[HTML]{9C5700} Turnigy SK8-ESC}
        & \cellcolor[HTML]{FFEB9C}{\color[HTML]{9C5700} AS}
        & \cellcolor[HTML]{FFC7CE}{\color[HTML]{9C0006} \xmark}
        & \cellcolor[HTML]{FFC7CE}{\color[HTML]{9C0006} \xmark}
        & \cellcolor[HTML]{C6EFCE}{\color[HTML]{006100} \cmark}
        & \cellcolor[HTML]{FFC7CE}{\color[HTML]{9C0006} \xmark}
        & \cellcolor[HTML]{FFC7CE}{\color[HTML]{9C0006} \xmark}
        & \cellcolor[HTML]{FFC7CE}{\color[HTML]{9C0006} \xmark}
        & \cellcolor[HTML]{C6EFCE}{\color[HTML]{006100} \cmark}
        & \cellcolor[HTML]{FFC7CE}{\color[HTML]{9C0006} \xmark}
        & \cellcolor[HTML]{FFC7CE}{\color[HTML]{9C0006} \xmark}
        & \cellcolor[HTML]{FFC7CE}{\color[HTML]{9C0006} \xmark}
        & \cellcolor[HTML]{FFC7CE}{\color[HTML]{9C0006} \xmark}
        \\
        BARC \cite{BARC2021}
        & \cellcolor[HTML]{C6EFCE}{\color[HTML]{006100} 1:10}
        & \cellcolor[HTML]{FFEB9C}{\color[HTML]{9C5700} \smark}
        & \cellcolor[HTML]{C6EFCE}{\color[HTML]{006100} \cmark}
        & \cellcolor[HTML]{FFEB9C}{\color[HTML]{9C5700} \$1,030}
        & \cellcolor[HTML]{FFC7CE}{\color[HTML]{9C0006} \xmark}
        & \cellcolor[HTML]{FFC7CE}{\color[HTML]{9C0006} \xmark}
        & \cellcolor[HTML]{C6EFCE}{\color[HTML]{006100} \cmark}
        & \cellcolor[HTML]{FFC7CE}{\color[HTML]{9C0006} \xmark}
        & \cellcolor[HTML]{C6EFCE}{\color[HTML]{006100} \cmark}
        & \cellcolor[HTML]{FFC7CE}{\color[HTML]{9C0006} \xmark}
        & \cellcolor[HTML]{FFC7CE}{\color[HTML]{9C0006} \xmark}
        & \cellcolor[HTML]{C6EFCE}{\color[HTML]{006100} \cmark}
        & \cellcolor[HTML]{FFC7CE}{\color[HTML]{9C0006} \xmark}
        & \cellcolor[HTML]{FFEB9C}{\color[HTML]{9C5700} ODROID-XU4}
        & \cellcolor[HTML]{C6EFCE}{\color[HTML]{006100} Arduino Nano}
        & \cellcolor[HTML]{FFEB9C}{\color[HTML]{9C5700} AS}
        & \cellcolor[HTML]{FFC7CE}{\color[HTML]{9C0006} \xmark}
        & \cellcolor[HTML]{FFC7CE}{\color[HTML]{9C0006} \xmark}
        & \cellcolor[HTML]{FFC7CE}{\color[HTML]{9C0006} \xmark}
        & \cellcolor[HTML]{FFC7CE}{\color[HTML]{9C0006} \xmark}
        & \cellcolor[HTML]{FFC7CE}{\color[HTML]{9C0006} \xmark}
        & \cellcolor[HTML]{FFC7CE}{\color[HTML]{9C0006} \xmark}
        & \cellcolor[HTML]{C6EFCE}{\color[HTML]{006100} \cmark}
        & \cellcolor[HTML]{FFC7CE}{\color[HTML]{9C0006} \xmark}
        & \cellcolor[HTML]{FFC7CE}{\color[HTML]{9C0006} \xmark}
        & \cellcolor[HTML]{FFC7CE}{\color[HTML]{9C0006} \xmark}
        & \cellcolor[HTML]{FFC7CE}{\color[HTML]{9C0006} \xmark}
        \\
        ORCA \cite{ORCA2021}
        & \cellcolor[HTML]{C6EFCE}{\color[HTML]{006100} 1:43}
        & \cellcolor[HTML]{FFEB9C}{\color[HTML]{9C5700} \smark}
        & \cellcolor[HTML]{C6EFCE}{\color[HTML]{006100} \cmark}
        & \cellcolor[HTML]{C6EFCE}{\color[HTML]{006100} \$960}
        & \cellcolor[HTML]{FFC7CE}{\color[HTML]{9C0006} \xmark}
        & \cellcolor[HTML]{FFC7CE}{\color[HTML]{9C0006} \xmark}
        & \cellcolor[HTML]{FFC7CE}{\color[HTML]{9C0006} \xmark}
        & \cellcolor[HTML]{FFC7CE}{\color[HTML]{9C0006} \xmark}
        & \cellcolor[HTML]{C6EFCE}{\color[HTML]{006100} \cmark}
        & \cellcolor[HTML]{FFC7CE}{\color[HTML]{9C0006} \xmark}
        & \cellcolor[HTML]{FFC7CE}{\color[HTML]{9C0006} \xmark}
        & \cellcolor[HTML]{FFC7CE}{\color[HTML]{9C0006} \xmark}
        & \cellcolor[HTML]{FFC7CE}{\color[HTML]{9C0006} \xmark}
        & \cellcolor[HTML]{FFC7CE}{\color[HTML]{9C0006} None}
        & \cellcolor[HTML]{C6EFCE}{\color[HTML]{006100} ARM Cortex M4}
        & \cellcolor[HTML]{FFEB9C}{\color[HTML]{9C5700} AS}
        & \cellcolor[HTML]{FFC7CE}{\color[HTML]{9C0006} \xmark}
        & \cellcolor[HTML]{FFC7CE}{\color[HTML]{9C0006} \xmark}
        & \cellcolor[HTML]{FFC7CE}{\color[HTML]{9C0006} \xmark}
        & \cellcolor[HTML]{C6EFCE}{\color[HTML]{006100} \cmark}
        & \cellcolor[HTML]{C6EFCE}{\color[HTML]{006100} \cmark}
        & \cellcolor[HTML]{FFC7CE}{\color[HTML]{9C0006} \xmark}
        & \cellcolor[HTML]{FFC7CE}{\color[HTML]{9C0006} \xmark}
        & \cellcolor[HTML]{FFC7CE}{\color[HTML]{9C0006} \xmark}
        & \cellcolor[HTML]{FFC7CE}{\color[HTML]{9C0006} \xmark}
        & \cellcolor[HTML]{FFEB9C}{\color[HTML]{9C5700} \smark}
        & \cellcolor[HTML]{FFC7CE}{\color[HTML]{9C0006} \xmark}
        \\
        HyphaROS \cite{HyphaROS-Racecar2021}
        & \cellcolor[HTML]{C6EFCE}{\color[HTML]{006100} 1:10}
        & \cellcolor[HTML]{FFEB9C}{\color[HTML]{9C5700} \smark}
        & \cellcolor[HTML]{C6EFCE}{\color[HTML]{006100} \cmark}
        & \cellcolor[HTML]{C6EFCE}{\color[HTML]{006100} \$600}
        & \cellcolor[HTML]{FFC7CE}{\color[HTML]{9C0006} \xmark}
        & \cellcolor[HTML]{FFC7CE}{\color[HTML]{9C0006} \xmark}
        & \cellcolor[HTML]{FFC7CE}{\color[HTML]{9C0006} \xmark}
        & \cellcolor[HTML]{FFC7CE}{\color[HTML]{9C0006} \xmark}
        & \cellcolor[HTML]{C6EFCE}{\color[HTML]{006100} \cmark}
        & \cellcolor[HTML]{FFC7CE}{\color[HTML]{9C0006} \xmark}
        & \cellcolor[HTML]{C6EFCE}{\color[HTML]{006100} \cmark}
        & \cellcolor[HTML]{FFC7CE}{\color[HTML]{9C0006} \xmark}
        & \cellcolor[HTML]{FFC7CE}{\color[HTML]{9C0006} \xmark}
        & \cellcolor[HTML]{FFEB9C}{\color[HTML]{9C5700} ODROID-XU4}
        & \cellcolor[HTML]{FFEB9C}{\color[HTML]{9C5700} ESC TBLE-02S}
        & \cellcolor[HTML]{FFEB9C}{\color[HTML]{9C5700} AS}
        & \cellcolor[HTML]{FFC7CE}{\color[HTML]{9C0006} \xmark}
        & \cellcolor[HTML]{FFC7CE}{\color[HTML]{9C0006} \xmark}
        & \cellcolor[HTML]{FFC7CE}{\color[HTML]{9C0006} \xmark}
        & \cellcolor[HTML]{FFC7CE}{\color[HTML]{9C0006} \xmark}
        & \cellcolor[HTML]{FFC7CE}{\color[HTML]{9C0006} \xmark}
        & \cellcolor[HTML]{FFC7CE}{\color[HTML]{9C0006} \xmark}
        & \cellcolor[HTML]{C6EFCE}{\color[HTML]{006100} \cmark}
        & \cellcolor[HTML]{FFC7CE}{\color[HTML]{9C0006} \xmark}
        & \cellcolor[HTML]{FFC7CE}{\color[HTML]{9C0006} \xmark}
        & \cellcolor[HTML]{FFC7CE}{\color[HTML]{9C0006} \xmark}
        & \cellcolor[HTML]{FFC7CE}{\color[HTML]{9C0006} \xmark}
        \\
        Donkey Car \cite{DonkeyCar2021}
        & \cellcolor[HTML]{C6EFCE}{\color[HTML]{006100} 1:16}
        & \cellcolor[HTML]{FFEB9C}{\color[HTML]{9C5700} \smark}
        & \cellcolor[HTML]{C6EFCE}{\color[HTML]{006100} \cmark}
        & \cellcolor[HTML]{C6EFCE}{\color[HTML]{006100} \$370}
        & \cellcolor[HTML]{FFC7CE}{\color[HTML]{9C0006} \xmark}
        & \cellcolor[HTML]{FFC7CE}{\color[HTML]{9C0006} \xmark}
        & \cellcolor[HTML]{FFC7CE}{\color[HTML]{9C0006} \xmark}
        & \cellcolor[HTML]{FFC7CE}{\color[HTML]{9C0006} \xmark}
        & \cellcolor[HTML]{FFC7CE}{\color[HTML]{9C0006} \xmark}
        & \cellcolor[HTML]{FFC7CE}{\color[HTML]{9C0006} \xmark}
        & \cellcolor[HTML]{FFC7CE}{\color[HTML]{9C0006} \xmark}
        & \cellcolor[HTML]{C6EFCE}{\color[HTML]{006100} \cmark}
        & \cellcolor[HTML]{FFC7CE}{\color[HTML]{9C0006} \xmark}
        & \cellcolor[HTML]{FFEB9C}{\color[HTML]{9C5700} Raspberry Pi}
        & \cellcolor[HTML]{FFEB9C}{\color[HTML]{9C5700} ESC}
        & \cellcolor[HTML]{FFEB9C}{\color[HTML]{9C5700} AS}
        & \cellcolor[HTML]{FFC7CE}{\color[HTML]{9C0006} \xmark}
        & \cellcolor[HTML]{FFC7CE}{\color[HTML]{9C0006} \xmark}
        & \cellcolor[HTML]{FFC7CE}{\color[HTML]{9C0006} \xmark}
        & \cellcolor[HTML]{FFC7CE}{\color[HTML]{9C0006} \xmark}
        & \cellcolor[HTML]{FFC7CE}{\color[HTML]{9C0006} \xmark}
        & \cellcolor[HTML]{C6EFCE}{\color[HTML]{006100} \cmark}
        & \cellcolor[HTML]{FFC7CE}{\color[HTML]{9C0006} \xmark}
        & \cellcolor[HTML]{FFC7CE}{\color[HTML]{9C0006} \xmark}
        & \cellcolor[HTML]{FFC7CE}{\color[HTML]{9C0006} \xmark}
        & \cellcolor[HTML]{FFC7CE}{\color[HTML]{9C0006} \xmark}
        & \cellcolor[HTML]{FFC7CE}{\color[HTML]{9C0006} \xmark}
        \\
        QCar \cite{QCar2021}
        & \cellcolor[HTML]{C6EFCE}{\color[HTML]{006100} 1:10}
        & \cellcolor[HTML]{FFC7CE}{\color[HTML]{9C0006} \xmark}
        & \cellcolor[HTML]{FFC7CE}{\color[HTML]{9C0006} \xmark}
        & \cellcolor[HTML]{FFC7CE}{\color[HTML]{9C0006} \$20,000}
        & \cellcolor[HTML]{FFC7CE}{\color[HTML]{9C0006} \xmark}
        & \cellcolor[HTML]{FFC7CE}{\color[HTML]{9C0006} \xmark}
        & \cellcolor[HTML]{C6EFCE}{\color[HTML]{006100} \cmark}
        & \cellcolor[HTML]{FFC7CE}{\color[HTML]{9C0006} \xmark}
        & \cellcolor[HTML]{C6EFCE}{\color[HTML]{006100} \cmark}
        & \cellcolor[HTML]{C6EFCE}{\color[HTML]{006100} \cmark}
        & \cellcolor[HTML]{C6EFCE}{\color[HTML]{006100} \cmark}
        & \cellcolor[HTML]{C6EFCE}{\color[HTML]{006100} \cmark}
        & \cellcolor[HTML]{C6EFCE}{\color[HTML]{006100} \cmark}
        & \cellcolor[HTML]{C6EFCE}{\color[HTML]{006100} Jetson TX2}
        & \cellcolor[HTML]{FFC7CE}{\color[HTML]{9C0006} Proprietary} 
        & \cellcolor[HTML]{FFEB9C}{\color[HTML]{9C5700} AS}
        & \cellcolor[HTML]{C6EFCE}{\color[HTML]{006100} \cmark}
        & \cellcolor[HTML]{C6EFCE}{\color[HTML]{006100} \cmark}
        & \cellcolor[HTML]{C6EFCE}{\color[HTML]{006100} \cmark}
        & \cellcolor[HTML]{FFC7CE}{\color[HTML]{9C0006} \xmark}
        & \cellcolor[HTML]{FFEB9C}{\color[HTML]{9C5700} \smark}
        & \cellcolor[HTML]{FFEB9C}{\color[HTML]{9C5700} \smark}
        & \cellcolor[HTML]{FFEB9C}{\color[HTML]{9C5700} \smark}
        & \cellcolor[HTML]{FFC7CE}{\color[HTML]{9C0006} \xmark}
        & \cellcolor[HTML]{FFC7CE}{\color[HTML]{9C0006} \xmark}
        & \cellcolor[HTML]{C6EFCE}{\color[HTML]{006100} \cmark}
        & \cellcolor[HTML]{FFC7CE}{\color[HTML]{9C0006} \xmark}
        \\
        DeepRacer \cite{DeepRacer2021}
        & \cellcolor[HTML]{C6EFCE}{\color[HTML]{006100} 1:18}
        & \cellcolor[HTML]{FFC7CE}{\color[HTML]{9C0006} \xmark}
        & \cellcolor[HTML]{FFC7CE}{\color[HTML]{9C0006} \xmark}
        & \cellcolor[HTML]{C6EFCE}{\color[HTML]{006100} \$400}
        & \cellcolor[HTML]{FFC7CE}{\color[HTML]{9C0006} \xmark}
        & \cellcolor[HTML]{FFC7CE}{\color[HTML]{9C0006} \xmark}
        & \cellcolor[HTML]{FFC7CE}{\color[HTML]{9C0006} \xmark}
        & \cellcolor[HTML]{FFC7CE}{\color[HTML]{9C0006} \xmark}
        & \cellcolor[HTML]{C6EFCE}{\color[HTML]{006100} \cmark}
        & \cellcolor[HTML]{FFC7CE}{\color[HTML]{9C0006} \xmark}
        & \cellcolor[HTML]{FFEB9C}{\color[HTML]{9C5700} \smark}
        & \cellcolor[HTML]{C6EFCE}{\color[HTML]{006100} \cmark}
        & \cellcolor[HTML]{FFC7CE}{\color[HTML]{9C0006} \xmark}
        & \cellcolor[HTML]{FFC7CE}{\color[HTML]{9C0006} Proprietary}
        & \cellcolor[HTML]{FFC7CE}{\color[HTML]{9C0006} Proprietary}
        & \cellcolor[HTML]{FFEB9C}{\color[HTML]{9C5700} AS}
        & \cellcolor[HTML]{FFC7CE}{\color[HTML]{9C0006} \xmark}
        & \cellcolor[HTML]{FFC7CE}{\color[HTML]{9C0006} \xmark}
        & \cellcolor[HTML]{FFC7CE}{\color[HTML]{9C0006} \xmark}
        & \cellcolor[HTML]{FFC7CE}{\color[HTML]{9C0006} \xmark}
        & \cellcolor[HTML]{FFC7CE}{\color[HTML]{9C0006} \xmark}
        & \cellcolor[HTML]{FFC7CE}{\color[HTML]{9C0006} \xmark}
        & \cellcolor[HTML]{FFC7CE}{\color[HTML]{9C0006} \xmark}
        & \cellcolor[HTML]{FFC7CE}{\color[HTML]{9C0006} \xmark}
        & \cellcolor[HTML]{FFC7CE}{\color[HTML]{9C0006} \xmark}
        & \cellcolor[HTML]{FFC7CE}{\color[HTML]{9C0006} \xmark}
        & \cellcolor[HTML]{C6EFCE}{\color[HTML]{006100} \cmark}
        \\
        Duckiebot \cite{Duckietown2017}
        & \cellcolor[HTML]{FFC7CE}{\color[HTML]{9C0006} N/A}
        & \cellcolor[HTML]{C6EFCE}{\color[HTML]{006100} \cmark}
        & \cellcolor[HTML]{C6EFCE}{\color[HTML]{006100} \cmark}
        & \cellcolor[HTML]{C6EFCE}{\color[HTML]{006100} \$450}
        & \cellcolor[HTML]{FFC7CE}{\color[HTML]{9C0006} \xmark}
        & \cellcolor[HTML]{FFC7CE}{\color[HTML]{9C0006} \xmark}
        & \cellcolor[HTML]{FFEB9C}{\color[HTML]{9C5700} \smark}
        & \cellcolor[HTML]{FFC7CE}{\color[HTML]{9C0006} \xmark}
        & \cellcolor[HTML]{FFEB9C}{\color[HTML]{9C5700} \smark}
        & \cellcolor[HTML]{FFC7CE}{\color[HTML]{9C0006} \xmark}
        & \cellcolor[HTML]{FFC7CE}{\color[HTML]{9C0006} \xmark}
        & \cellcolor[HTML]{C6EFCE}{\color[HTML]{006100} \cmark}
        & \cellcolor[HTML]{FFC7CE}{\color[HTML]{9C0006} \xmark}
        & \cellcolor[HTML]{FFEB9C}{\color[HTML]{9C5700} Raspberry Pi}
        & \cellcolor[HTML]{FFC7CE}{\color[HTML]{9C0006} None}
        & \cellcolor[HTML]{FFC7CE}{\color[HTML]{9C0006} DD}
        & \cellcolor[HTML]{FFEB9C}{\color[HTML]{9C5700} \smark}
        & \cellcolor[HTML]{FFC7CE}{\color[HTML]{9C0006} \xmark}
        & \cellcolor[HTML]{FFEB9C}{\color[HTML]{9C5700} \smark}
        & \cellcolor[HTML]{FFEB9C}{\color[HTML]{9C5700} \smark}
        & \cellcolor[HTML]{FFC7CE}{\color[HTML]{9C0006} \xmark}
        & \cellcolor[HTML]{FFC7CE}{\color[HTML]{9C0006} \xmark}
        & \cellcolor[HTML]{C6EFCE}{\color[HTML]{006100} \cmark}
        & \cellcolor[HTML]{FFC7CE}{\color[HTML]{9C0006} \xmark}
        & \cellcolor[HTML]{FFC7CE}{\color[HTML]{9C0006} \xmark}
        & \cellcolor[HTML]{FFC7CE}{\color[HTML]{9C0006} \xmark}
        & \cellcolor[HTML]{FFC7CE}{\color[HTML]{9C0006} \xmark}
        \\
        TurtleBot3 \cite{Turtlebot2021}
        & \cellcolor[HTML]{FFC7CE}{\color[HTML]{9C0006} N/A}
        & \cellcolor[HTML]{C6EFCE}{\color[HTML]{006100} \cmark}
        & \cellcolor[HTML]{C6EFCE}{\color[HTML]{006100} \cmark}
        & \cellcolor[HTML]{C6EFCE}{\color[HTML]{006100} \$590}
        & \cellcolor[HTML]{FFC7CE}{\color[HTML]{9C0006} \xmark}
        & \cellcolor[HTML]{FFC7CE}{\color[HTML]{9C0006} \xmark}
        & \cellcolor[HTML]{C6EFCE}{\color[HTML]{006100} \cmark}
        & \cellcolor[HTML]{FFC7CE}{\color[HTML]{9C0006} \xmark}
        & \cellcolor[HTML]{C6EFCE}{\color[HTML]{006100} \cmark}
        & \cellcolor[HTML]{FFC7CE}{\color[HTML]{9C0006} \xmark}
        & \cellcolor[HTML]{C6EFCE}{\color[HTML]{006100} \cmark}
        & \cellcolor[HTML]{FFEB9C}{\color[HTML]{9C5700} \smark}
        & \cellcolor[HTML]{FFC7CE}{\color[HTML]{9C0006} \xmark}
        & \cellcolor[HTML]{FFEB9C}{\color[HTML]{9C5700} Raspberry Pi}
        & \cellcolor[HTML]{C6EFCE}{\color[HTML]{006100} OpenCR}
        & \cellcolor[HTML]{FFC7CE}{\color[HTML]{9C0006} DD}
        & \cellcolor[HTML]{FFC7CE}{\color[HTML]{9C0006} \xmark}
        & \cellcolor[HTML]{FFC7CE}{\color[HTML]{9C0006} \xmark}
        & \cellcolor[HTML]{FFEB9C}{\color[HTML]{9C5700} \smark}
        & \cellcolor[HTML]{FFC7CE}{\color[HTML]{9C0006} \xmark}
        & \cellcolor[HTML]{FFC7CE}{\color[HTML]{9C0006} \xmark}
        & \cellcolor[HTML]{FFC7CE}{\color[HTML]{9C0006} \xmark}
        & \cellcolor[HTML]{C6EFCE}{\color[HTML]{006100} \cmark}
        & \cellcolor[HTML]{FFEB9C}{\color[HTML]{9C5700} \smark}
        & \cellcolor[HTML]{FFC7CE}{\color[HTML]{9C0006} \xmark}
        & \cellcolor[HTML]{FFC7CE}{\color[HTML]{9C0006} \xmark}
        & \cellcolor[HTML]{FFC7CE}{\color[HTML]{9C0006} \xmark}
        \\
        \multicolumn{24}{l}{}
        \\
        \multicolumn{28}{l}{\normalsize$^{\mathrm{\dagger}}$All costs are ceiled to the nearest \$10. $^{\mathrm{\ddagger}}$Actuation mechanisms comprise Ackermann-steered (AS), differential-drive (DD), and 4-wheel-drive 4-wheel-steer (4WD4WS) configurations.}
        \\
        \multicolumn{28}{l}{\normalsize\color[HTML]{006100} \cmark \color{black} $\!$ indicates complete fulfillment; \color[HTML]{9C5700} \smark \color{black} $\!$ indicates conditional, unsupported or partial fulfillment; and \color[HTML]{9C0006} \xmark \color{black} $\!$ indicates non-fulfillment.}
        \\
    \end{tabular}
}
\end{table*}

In terms of developing scaled autonomous vehicles, academic institutions have contributed platforms like \cite{MIT-Racecar2017, AutoRally2021, F1TENTH2019, MuSHR2019, ORCA2021, DSV2017, BARC2021}. Community-driven projects such as \cite{HyphaROS-Racecar2021, DonkeyCar2021} have also emerged, which are often tailored for specific applications. Moreover, commercial products like \cite{QCar2021, DeepRacer2021} have also entered the market, but their closed-source nature and prohibitive costs limit their accessibility to the broader community. Additionally, scaled robot platforms like \cite{Turtlebot2021, Duckietown2017} remain valuable for teaching fundamental autonomy concepts. However, none of the aforementioned platforms (refer Table \ref{tab1}) contributes primarily towards novel vehicular configuration or architecture. Other works such as \cite{electronics12163511, s22062144} have recently prototyped over-actuated scaled vehicles for validating their control algorithms, but none of them are open-sourced. Additionally, these platforms lack comprehensive autonomy features with only the latter offering truly independent driving and steering capability, although at a much larger scale (1:5) and without extended steering limits. To the best of authors’ knowledge, Nigel (refer Fig. \ref{fig2}) is the first open-source\footnote{GitHub: \texttt{\url{https://github.com/AutoDRIVE-Ecosystem}}} mechatronically redundant \cite{AutoDRIVEMechatronics2023} autonomous vehicle platform offering comprehensive autonomy features as well as independent all-wheel driving and independent all-wheel steering (i.e., independent 4WD4WS) configuration with extended ($\pm90^\circ$) steering angles, within a small footprint of 1:14 scale. Additionally, Nigel is a part of the larger AutoDRIVE Ecosystem\footnote{Website: \texttt{\url{https://autodrive-ecosystem.github.io}}} \cite{AutoDRIVEEcosystem2022, AutoDRIVEReport2021}, which also offers a high-fidelity digital-twin simulation platform \cite{AutoDRIVESimulator2021, AutoDRIVESimulatorReport2020}, as well as flexible application programming interfaces (APIs) to develop low and high-level autonomy algorithms.

In the context of developing control strategies for vehicles with unconventional architectures, prior works have applied techniques such as inverse dynamics control \cite{Zhang2021}, coordinated motion control \cite{electronics11223731}, adaptive steering control \cite{10.1115/DSCC2012-MOVIC2012-8603} and model predictive control \cite{SCHWARTZ2019162, app8061000}. Although some of these works discuss the adaptability and robustness of the designed controllers in simulation, they fail to guarantee similar performance in real-world conditions. Other recent works such as \cite{1081954} and \cite{LI2024104621} have applied cascaded feedforward control and robust $H_\infty$ control, respectively, to handle environmental uncertainties and disturbances using over-actuated vehicle architectures. Similar robust control techniques have also been applied to automotive systems for yaw-plane \cite{221348, doi:10.1080/00423114.2013.879190, 8961203}, roll-plane \cite{s21237850} or vertical dynamics stabilization \cite{Babawuro_2020}, emergency braking \cite{5160361}, trajectory tracking \cite{9108078}, etc. However, mixed $H_2$-$H_\infty$ robust optimal control has not been explored in the automotive field, especially for 4WD4WS autonomous vehicles, and our work seeks to fill in this void.

Specifically, our work seeks to develop a generalizable framework for reliably bridging the sim2real gap by applying the rich theory of multi-model multi-objective control, which seeks the Pareto-optimal solution of $H_2$ and $H_\infty$ performance tradeoff with $D$-stability guarantee. It is to be noted that, instead of bridging the sim2real gap by increasing simulation fidelity \cite{SAMAK2023277}, this work intentionally widens the reality gap by using a linearized dynamics model to synthesize the robust optimal controller, which is hypothesized to bridge the dynamics gap during real-world deployment. To this end, we benchmark and validate robust stabilizing, as well as steady-state tracking control of Nigel with seamless sim2real transfer.

To recapitulate, our contributions are itemized below:
\begin{itemize}
    \item Completely open-source design architecture of a novel 1:14 scale independent 4WD4WS autonomous vehicle with extended steering limits is presented.
    \item Non-linear dynamics model of such a novel vehicle architecture is derived and transformed into a linear uncertain system model with reasonable parameter identification.
    \item Robust sim2real control framework is established and validated through exhaustive experimentation and benchmarking in simulation as well as real-world settings.
\end{itemize}



\section{Vehicle Design Architecture}
\label{Section: Vehicle Design Architecture}

\begin{figure}[t]
	\centering
	\includegraphics[width=\linewidth]{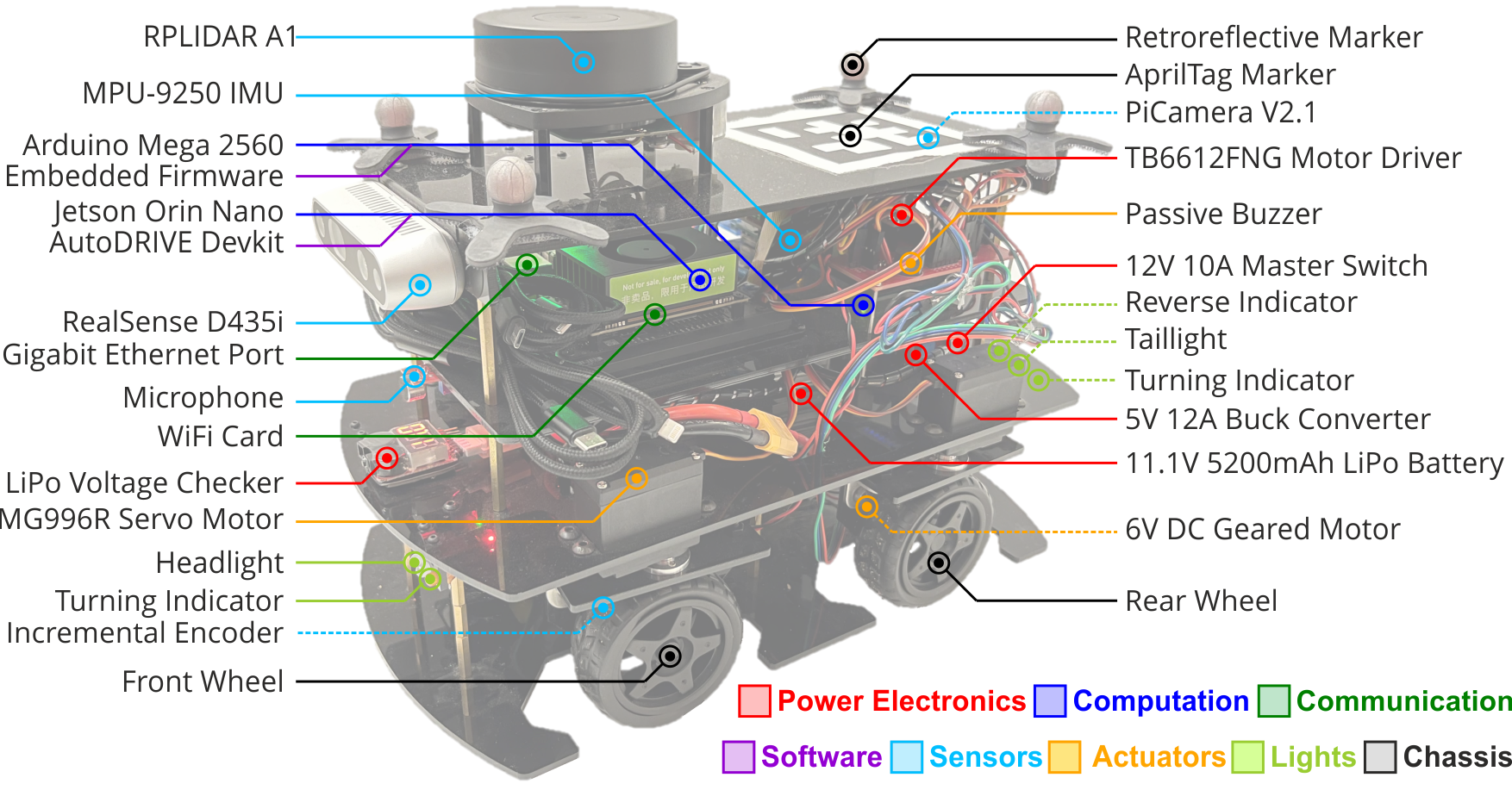}
	\caption{Various components and sub-systems of Nigel. Dashed lines indicate occluded or hidden components. A video highlighting key features of Nigel is available at \texttt{\url{https://youtu.be/UVIShZuZmpg}}}
	\label{fig2}
\end{figure}

The key design objective for vehicle architecture discussed in this work was to develop a mechatronically redundant scaled autonomous vehicle. The resulting prototype, Nigel, offers redundant driving and steering actuation, a comprehensive sensor suite, high-performance computational resources, and a standard vehicular signaling system (refer Fig. \ref{fig2}).

\textit{\textbf{Chassis:}}
Nigel is a 1:14 scale autonomous vehicle comprising four modular platforms, each housing distinct components of the vehicle.

\textit{\textbf{Power Electronics:}}
Nigel is powered using an 11.1 V 5200 mAh lithium-polymer (LiPo) battery, whose health is monitored by a voltage checker. Other components such as the master switch, buck converter and motor drivers help route the power to appropriate sub-systems of the vehicle.

\textit{\textbf{Sensor Suite:}}
Nigel hosts a comprehensive sensor suite comprising throttle/steering feedbacks and 1920 CPR incremental encoders for all 4 wheels, a microphone, a 3-axis indoor-positioning system (IPS) using retroreflective/fiducial markers, a 9-axis inertial measurement unit (IMU) with raw/filtered measurements, multiple RGB/RGB-D/stereo camera(s) in the front/rear, and a 360$^\circ$ FOV planar LIDAR.

\textit{\textbf{Computation, Communication and Software:}}
Nigel adopts Jetson Orin Nano Developer Kit for most of its high-level autonomy algorithms and V2X communication and hosts Arduino Mega for acquiring and filtering raw sensor data, and controlling the actuators/lights/indicators.

\begin{figure*}[b]
	\centering
	\includegraphics[width=\linewidth]{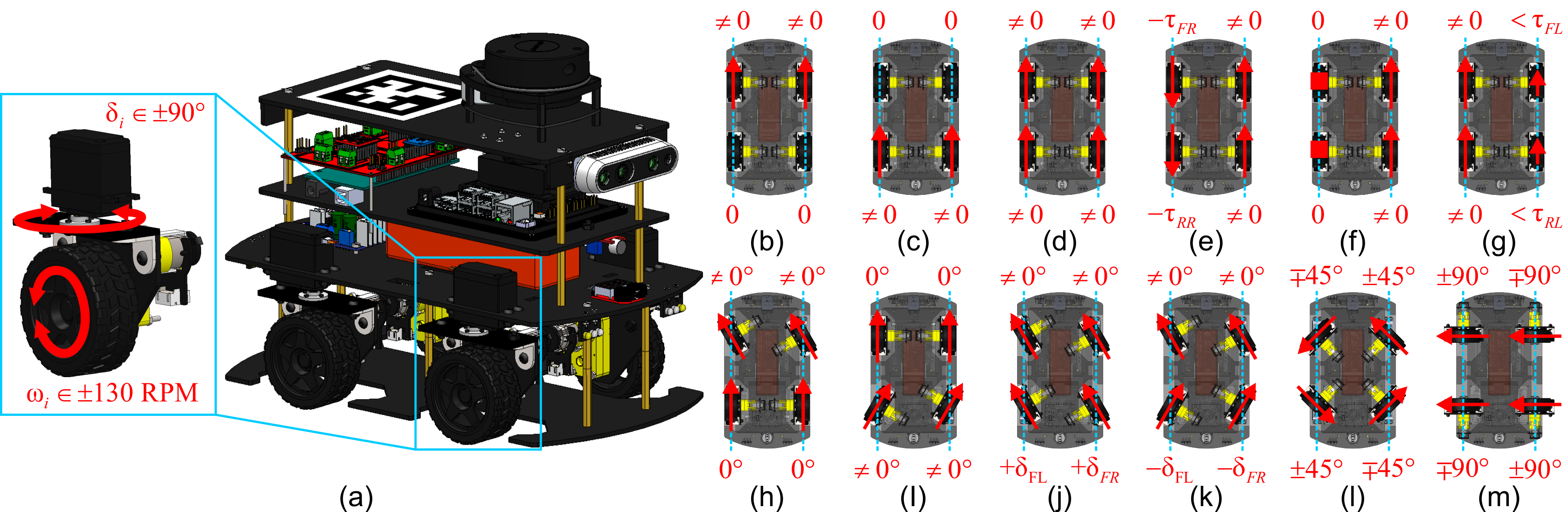}
	\caption{Independent 4WD4WS architecture of Nigel: (a) denotes 2-DOF actuation redundancy per wheel; (b)-(g) depict common drive configurations including front-wheel drive, rear-wheel drive, all-wheel drive, neutral-steer drive, pivot-steer drive and torque vectoring drive; and (h)-(m) depict common steering configurations including front-wheel steer, rear-wheel steer, all-wheel in-phase steer, all-wheel out-of-phase steer, oblique steer and crab-walk steer.}
	\label{fig3}
\end{figure*}

\textit{\textbf{Actuators:}}
Nigel is actuated using four 6V 160 RPM rated 120:1 DC geared motors to drive its wheels, and four 9.4 kgf.cm servo motors to steer them; the steering actuators are saturated at $\pm$ 90$^\circ$ w.r.t. zero-steer value (see Fig. \ref{fig3}). All the actuators are operated at 5V, which translates to a maximum speed of $\sim$130 RPM for driving and $\sim$0.19 s/60$^\circ$ for steering. The steering actuators are positioned directly above the respective tire contact patch, which enables zero camber gain and keeps the actuator effort stable.

\textit{\textbf{Lights and Indicators:}}
Nigel's lighting system comprises dual-mode headlights, triple-mode turning indicators, and automated taillights with reverse indicators. Additionally, Nigel is also provided with a buzzer to allow acoustic indication.

Kinematic analysis of Nigel's configuration reveals that it has $\delta_M = \delta_m + \delta_s = 1 + 2 = 3$ degrees of maneuverability; this is superior among all passively stable configurations possible. Elucidation follows. Considering the vehicle's configuration space $\mathscr{C} \in \mathbb{R}^m$, the $n$-dimensional admissible velocity space, which is a sub-space of the generalized velocity space $\mathscr{V} \in \mathbb{R}^m$ (composed of the time derivatives of the generalized coordinates of $\mathscr{C}$), governs the vehicle's differential degrees of freedom (a.k.a. degree of mobility, $\delta_m = n$). Given the sliding constraint matrix $\mathbf{C_1(\delta_i)} = \begin{bmatrix} \mathbf{C_{1_f}}\\ \mathbf{C_{1_s}(\delta_i)}\end{bmatrix}$ for fixed ($f$) and steerable ($s$) wheels, this translates to $\delta_m$ being the dimension of the right null space of $\mathbf{C_1(\delta_i)}$, i.e., $\delta_m = \mathrm{dim\;} N[\mathbf{C_1(\delta_i)}] = 3 - \mathrm{rank}[\mathbf{C_1(\delta_i)}]$. The degree of steerability, $\delta_s$, is governed by the sliding constraints imposed by the $i$ steerable wheels, with $\delta_s = \mathrm{rank}[\mathbf{C_{1_s}(\delta_i)}]$; $0 \leq \delta_s \leq 2$.


\section{Dynamics Modeling and Analysis}
\label{Section: Dynamics Modeling and Analysis}

\begin{figure}[t]
	\centering
	\includegraphics[width=\linewidth]{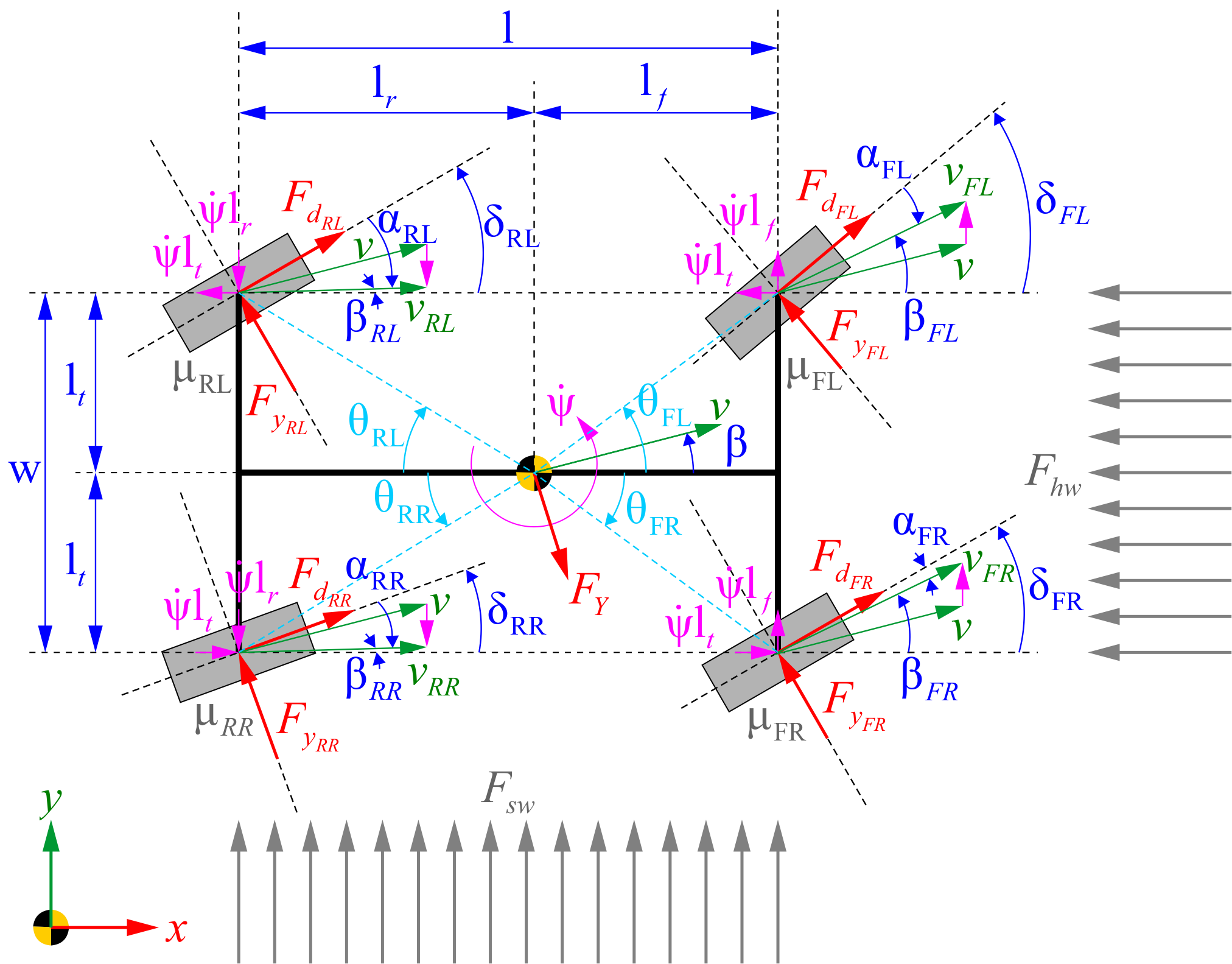}
	\caption{Vehicle dynamics model for an independent 4WD4WS vehicle.}
	\label{fig4}
\end{figure}

Considering the notations presented in Fig. \ref{fig4}, and extending the seminal works \cite{milliken1995race, rajamani2011vehicle}, we can deduce the non-linear yaw-plane vehicle dynamics model for an independent 4WD4WS configuration as given in Eq. \ref{eq_2_1a}-\ref{eq_2_1c}.

\footnotesize
\begin{subequations}
\begin{equation}
\begin{split}
    \mathbf{\sum F_x: \:} m \left(\dot{v}\cos(\beta) - v \dot{\beta}\sin(\beta) - \dot{\psi}v\sin(\beta) \right) = \left[ \frac{\tau_{RL}}{r_{RL}} \cos(\delta_{RL}) \right. \\
    \left. + \frac{\tau_{RR}}{r_{RR}} \cos(\delta_{RR})
     + \frac{\tau_{FL}}{r_{FL}} \cos(\delta_{FL}) + \frac{\tau_{FR}}{r_{FR}} \cos(\delta_{FR}) \right] - \left\{ \mu_{RL} C_{RL} \right. \\
     \left. \left[ \delta_{RL} - \tan^{-1} \left( \frac{\dot{y} - \dot{\psi}l_r}{\dot{x} - \dot{\psi}l_t} \right) \right] \sin(\delta_{RL}) + \mu_{RR} C_{RR} \left[ \delta_{RR} - \tan^{-1} \right. \right. \\
     \left. \left( \frac{\dot{y} - \dot{\psi}l_r}{\dot{x} + \dot{\psi}l_t} \right) \right] \sin(\delta_{RR}) + \left. \mu_{FL} C_{FL} \left[ \delta_{FL} - \tan^{-1} \left( \frac{\dot{y} + \dot{\psi}l_f}{\dot{x} - \dot{\psi}l_t} \right) \right] \right. \\
     \left. \sin(\delta_{FL}) + \mu_{FR} C_{FR} \left[ \delta_{FR} - \tan^{-1} \left( \frac{\dot{y} + \dot{\psi}l_f}{\dot{x} + \dot{\psi}l_t} \right) \right] \sin(\delta_{FR}) \right\} - F_{hw}
\end{split}
\label{eq_2_1a}
\end{equation}

\begin{equation}
\begin{split}
    \mathbf{\sum F_y: \:} m \left(\dot{v}\sin(\beta) + v \dot{\beta}\cos(\beta) - \dot{\psi}v\cos(\beta) \right) = \left[ \frac{\tau_{RL}}{r_{RL}} \sin(\delta_{RL}) \right. \\
    \left. + \frac{\tau_{RR}}{r_{RR}} \sin(\delta_{RR})
     + \frac{\tau_{FL}}{r_{FL}} \sin(\delta_{FL}) + \frac{\tau_{FR}}{r_{FR}} \sin(\delta_{FR}) \right] + \left\{ \mu_{RL} C_{RL} \right. \\
     \left. \left[ \delta_{RL} - \tan^{-1} \left( \frac{\dot{y} - \dot{\psi}l_r}{\dot{x} - \dot{\psi}l_t} \right) \right] \cos(\delta_{RL}) + \mu_{RR} C_{RR} \left[ \delta_{RR} - \tan^{-1} \right. \right. \\
     \left. \left( \frac{\dot{y} - \dot{\psi}l_r}{\dot{x} + \dot{\psi}l_t} \right) \right] \cos(\delta_{RR}) + \left. \mu_{FL} C_{FL} \left[ \delta_{FL} - \tan^{-1} \left( \frac{\dot{y} + \dot{\psi}l_f}{\dot{x} - \dot{\psi}l_t} \right) \right] \right. \\
     \left. \cos(\delta_{FL}) + \mu_{FR} C_{FR} \left[ \delta_{FR} - \tan^{-1} \left( \frac{\dot{y} + \dot{\psi}l_f}{\dot{x} + \dot{\psi}l_t} \right) \right] \cos(\delta_{FR}) \right\} + F_{sw}
\end{split}
\label{eq_2_1b}
\end{equation}

\begin{equation}
\begin{split}
    \mathbf{\sum M_z: \:} I_z \ddot{\psi} = l_f \left\{-\frac{\tau_{FL}}{r_{FL}} \sin(\delta_{FL} - \theta_{FL}) + \frac{\tau_{FR}}{r_{FR}} \sin(\delta_{FR} + \theta_{FR}) \right. \\
    \left. + \mu_{FL} C_{FL} \left[ \delta_{FL} - \tan^{-1} \left( \frac{\dot{y} + \dot{\psi}l_f}{\dot{x} - \dot{\psi}l_t} \right) \right] \cos(\delta_{FL} - \theta_{FL}) + \mu_{FR} C_{FR} \right. \\
    \left. \left[ \delta_{FR} - \tan^{-1} \left( \frac{\dot{y} + \dot{\psi}l_f}{\dot{x} + \dot{\psi}l_t} \right) \right] \cos(\delta_{FR} + \theta_{FR}) \right\} - l_r \left\{\frac{\tau_{RL}}{r_{RL}} \sin(\delta_{RL} \right. \\
    \left. + \theta_{RL}) - \frac{\tau_{RR}}{r_{RR}} \sin(\theta_{RR} - \delta_{RR}) + \mu_{RL} C_{RL} \left[ \delta_{RL} - \tan^{-1} \right. \right. \\ 
    \left. \left. \left( \frac{\dot{y} - \dot{\psi}l_r}{\dot{x} - \dot{\psi}l_t} \right) \right] \cos(\delta_{RL} + \theta_{RL}) + \mu_{RR} C_{RR} \left[ \delta_{RR} - \tan^{-1} \right. \right. \\
    \left. \left. \left( \frac{\dot{y} - \dot{\psi}l_r}{\dot{x} + \dot{\psi}l_t} \right) \right] \cos(\theta_{RR} - \delta_{RR}) \right\} + \left ( \frac{l_f - l_r}{2} \right ) F_{sw}
\end{split}
\label{eq_2_1c}
\end{equation}
\end{subequations}
\normalsize
Here, $m$ is the vehicle mass, $v$ is the vehicle velocity, $\beta = \tan^{-1}\left( \frac{\dot{y} \pm \dot{\psi}l_{f/r}}{\dot{x} \pm \dot{\psi}l_t} \right)$ is the vehicle slip angle, $\psi$ is the vehicle yaw angle, $\tau_i$ are the drive torques, $r_i$ are the wheel radii, $\delta_i$ are the steering angles, $\mu_i$ are the frictional coefficients for each road-tire interconnect, $C_i$ are the tire stiffness values, $\alpha_i = \delta_i - \beta$ are the tire slip angles, $l_f$ and $l_r$ are the front and rear center-of-gravity (CG) offsets, $I_z$ is the vehicle yaw moment of inertia, $\theta_i$ are the angular wheel locations w.r.t. CG, while $F_{hw}$ and $F_{sw}$ are respectively the head and side winds. Note that $F_{y_i} = \mu_i C_i \alpha_i$ and $F_{d_i} = \begin{cases} \tau_i/r_i; & \text{if } \tau_i/r_i \leq \mu_i N_i \\ \mu_i N_i;  & \text{otherwise} \end{cases}$ are tire forces and drive forces for $i$-th wheel, respectively.

The non-linear model in Eq. \ref{eq_2_1a}-\ref{eq_2_1c} can be linearized by applying the small-angle approximation so that $\sin(\odot)\approx\odot$, $\cos(\odot)\approx1$ and $\tan^{-1}(\odot)\approx\odot$. Additionally, we assume that the vehicle is driving at a constant velocity, implying $\dot{v} \approx 0 \Rightarrow \tau_i \approx 0$. Finally, it is also assumed that $\dot{x} \gg \dot{\psi}l_t \Rightarrow \dot{x} \pm \dot{\psi}l_t \approx \dot{x}$. Also note that $F_{sw}$ is hereafter referred to as $F_w$ for simplicity, since $F_{hw}$ does not affect the linearized model.

\footnotesize
\begin{equation}
\begin{bmatrix} \dot{\beta} \\ \ddot{\psi} \end{bmatrix} = \begin{bmatrix} a_{11} & a_{12} \\ a_{21} & a_{22} \end{bmatrix} \begin{bmatrix} \beta \\ \dot{\psi} \end{bmatrix} + \begin{bmatrix} b_{11} & b_{12} & b_{13} & b_{14} \\ b_{21} & b_{22} & b_{23} & b_{24} \end{bmatrix} \begin{bmatrix} \delta_{FL} \\ \delta_{FR} \\ \delta_{RL} \\ \delta_{RR} \end{bmatrix} + \begin{bmatrix} d_1 \\ d_2 \end{bmatrix} F_w
\label{eq_2_2}
\end{equation}
\normalsize
where
\footnotesize
\begin{equation*}
    \begin{aligned}
    a_{11} =  & \frac{-1}{mv} \left( \mu_{FL} C_{FL} + \mu_{FR} C_{FR} + \mu_{RL} C_{RL} + \mu_{RR} C_{RR} \right) \\
    a_{12} = & \left\{ \frac{1}{mv^2} \left[ l_r \left( \mu_{RL} C_{RL} + \mu_{RR} C_{RR}\right) - l_f \left(\mu_{FL} C_{FL} \right. \right. \right. \\
    & \left. \left. \left. + \mu_{FR} C_{FR} \right) \right] \right\} -1 \\
    a_{21} = & \frac{1}{I_z} \left[ l_r \left( \mu_{RL} C_{RL} + \mu_{RR} C_{RR}\right) - l_f \left(\mu_{FL} C_{FL} + \mu_{FR} C_{FR} \right) \right] \\
    a_{22} = & \frac{-1}{I_zv} \left[ l_f^2 \left( \mu_{FL} C_{FL} + \mu_{FR} C_{FR}\right) + l_r^2 \left(\mu_{RL} C_{RL} + \mu_{RR} C_{RR} \right) \right]
    \end{aligned}
\end{equation*}
\begin{equation*}
\begin{matrix*}[l]
    \begin{matrix*}[l]
    & b_{11} = \frac{\mu_{FL} C_{FL}}{mv}
    & b_{21} = \frac{l_f \mu_{FL} C_{FL}}{I_z}
    & b_{12} = \frac{\mu_{FR} C_{FR}}{mv} \\
    & b_{22} = \frac{l_f \mu_{FR} C_{FR}}{I_z}
    & b_{13} = \frac{\mu_{RL} C_{RL}}{mv}
    & b_{23} = \frac{-l_r \mu_{RL} C_{RL}}{I_z}
    \end{matrix*} \\
    \begin{matrix*}[l]
    & b_{14} = \frac{\mu_{RR} C_{RR}}{mv}
    & b_{24} = \frac{-l_r \mu_{RR} C_{RR}}{I_z}
    & d_{1} = \frac{1}{mv}
    & d_{2} = \frac{l_f-l_r}{2} 
    \end{matrix*}
\end{matrix*}
\end{equation*}
\normalsize

\begin{table}[t]
\centering
\caption{Measured and identified parameters of Nigel}
\label{tab2}
\resizebox{\columnwidth}{!}{%
\begin{tabular}{llll}
\hline
\multicolumn{1}{l|}{\textbf{Parameter}}      & \multicolumn{1}{l|}{\textbf{Symbol}} & \multicolumn{1}{l|}{\textbf{Value}} & \textbf{Unit} \\ \hline
\multicolumn{4}{l}{\textit{Measured Parameters}}                                                              \\ \hline
\multicolumn{1}{l|}{Mass}                  & \multicolumn{1}{l|}{$m$}   & \multicolumn{1}{l|}{2.68} & kg \\
\multicolumn{1}{l|}{Dimensions (L$\times$B$\times$H)}             & \multicolumn{1}{l|}{---}   & \multicolumn{1}{l|}{0.318$\times$0.175$\times$0.257} & m  \\
\multicolumn{1}{l|}{Wheelbase}             & \multicolumn{1}{l|}{$l$}   & \multicolumn{1}{l|}{0.14155} & m  \\
\multicolumn{1}{l|}{Track width}           & \multicolumn{1}{l|}{$l_t$} & \multicolumn{1}{l|}{0.14724} & m  \\
\multicolumn{1}{l|}{Wheel radius}          & \multicolumn{1}{l|}{$r_i$} & \multicolumn{1}{l|}{0.0325} & m  \\ \hline
\multicolumn{4}{l}{\textit{Identified Parameters}}                                                            \\ \hline
\multicolumn{1}{l|}{Yaw moment of inertia} & \multicolumn{1}{l|}{$I_z$}   & \multicolumn{1}{l|}{0.01944} &  kg.m$^{2}$  \\
\multicolumn{1}{l|}{CG front offset}       & \multicolumn{1}{l|}{$l_f$} & \multicolumn{1}{l|}{0.06226} & m  \\
\multicolumn{1}{l|}{CG rear offset}        & \multicolumn{1}{l|}{$l_r$} & \multicolumn{1}{l|}{0.07929} & m  \\
\multicolumn{1}{l|}{Friction coefficient}        & \multicolumn{1}{l|}{$\mu_i$} & \multicolumn{1}{l|}{0.4 (nominal)} & ---  \\
\multicolumn{1}{l|}{Tire stiffness} & \multicolumn{1}{l|}{$C_i$}           & \multicolumn{1}{l|}{22.4768}               & N.rad$^{-1}$         \\ \hline
\end{tabular}%
}
\end{table}

\begin{figure}[b]
  \centering
  \begin{tabular}{@{}c@{}}
    \includegraphics[width=.49\linewidth]{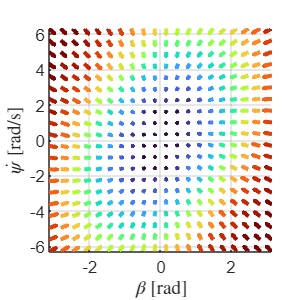} \\ \small (a)
  \end{tabular}
  \begin{tabular}{@{}c@{}}
    \includegraphics[width=.49\linewidth]{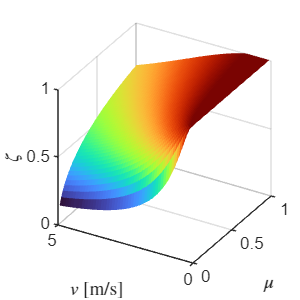} \\ \small (b)
  \end{tabular}
  \caption{Analysis of vehicle dynamics: (a) system phase portrait @ $\mu$ = 0.4, $v$ = 0.35 m/s; and (b) system damping ratio $\zeta$ as a function of $\mu$ and $v$.}
  \label{fig5}
\end{figure}

We analyzed the model obtained in Eq. \ref{eq_2_2} based on the parameters identified (refer Table \ref{tab2}) from lab experiments. The system has a stable equilibrium point under nominal conditions (refer Fig. \hyperref[fig5]{\ref*{fig5}(a)}), but drastic variation in operating conditions (friction or velocity) significantly affects the system's damping ratio, $\zeta$ = $\frac{-\lambda_R}{\sqrt{\lambda_R^2+\lambda_I^2}}$ (refer Fig. \hyperref[fig5]{\ref*{fig5}(b)}) where $\lambda = \lambda_R \pm \lambda_I \iota$ are eigenvalues of the system matrix. This follows that the system is prone to face sim2real transfer issues if the real-world conditions ($\mu_i$) are different from the simulation or uncertain in general, even at a constant velocity ($v$).

\section{Robust Sim2Real Control Framework}
\label{Section: Robust Sim2Real Control Framework}

\begin{figure}[ht]
    \centering
    \includegraphics[width=\linewidth]{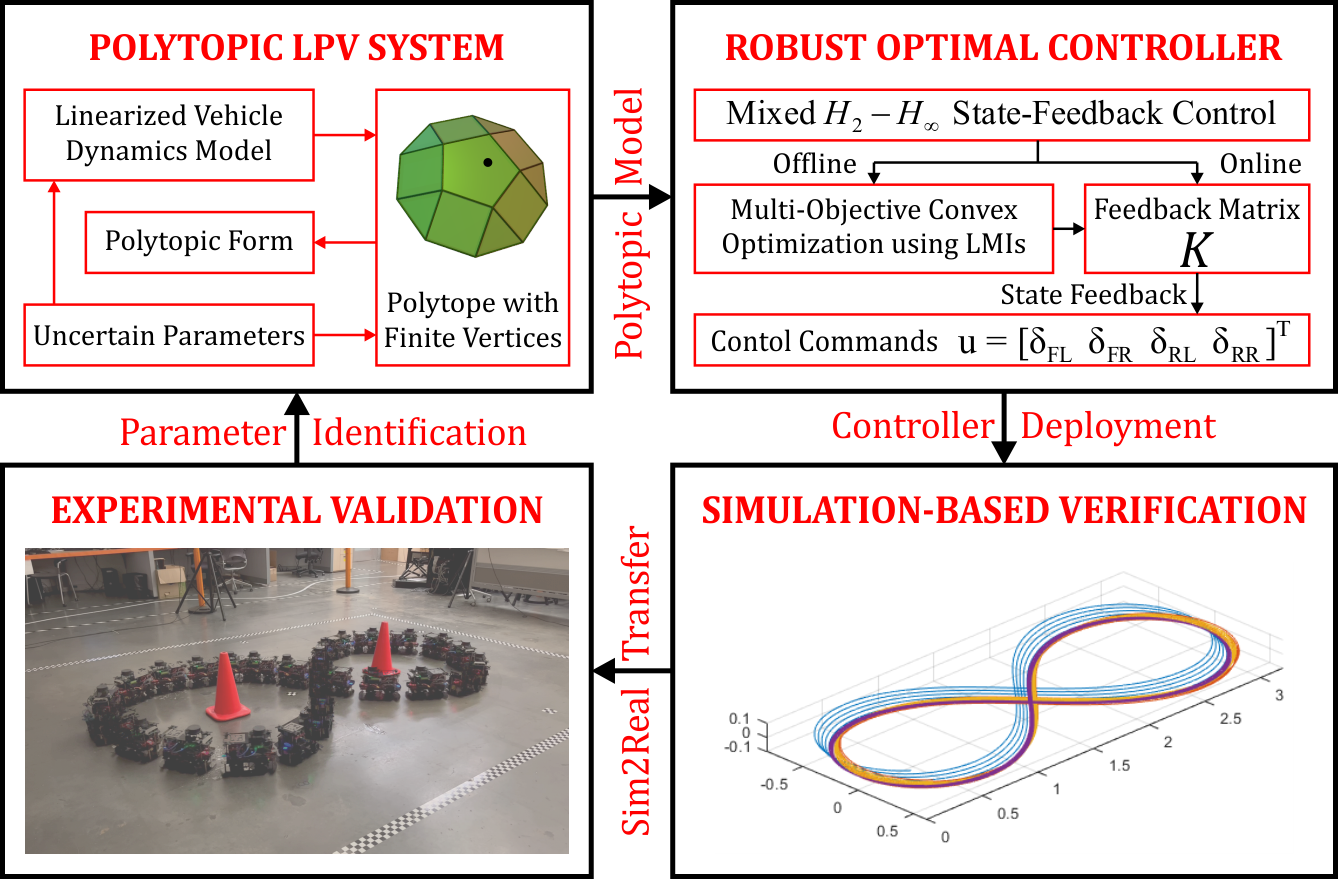}
    \caption{Structure of the presented robust sim2real control framework.}
    \label{fig6}
\end{figure}


The proposed framework (see Fig. \ref{fig6}) begins with formulating a linear parameter-varying (LPV) description of the system in polytopic form. Then, a robust optimal controller is synthesized by solving a semi-definite program using linear matrix inequalities (LMIs). Finally, the closed-loop system response is verified in simulation, followed by sim2real validation.

To this end, a generalized representation of the open-loop system $S_{ol}$ is obtained (refer Eq. \ref{eq_3_1}).
\begin{equation}
S_{ol}:=
\begin{cases}
\mathbf{\dot{x}_p = A_p x_p + B_p u + D_p w} \\
\mathbf{y_1 = C_{p_1} x_p + B_{y_1} u + D_y w} \\
\mathbf{y_2 = C_{p_2} x_p + B_{y_2} u} \\
\mathbf{z_p = M_p x_p + D_z w} \\
\end{cases}
\label{eq_3_1}
\end{equation}
Here, $\mathbf{u} = \begin{bmatrix} \delta_{FL} & \delta_{FR} & \delta_{RL} & \delta_{RR} \end{bmatrix}^T$ are the control inputs, $\mathbf{x_p} = \begin{bmatrix} \beta & \dot{\psi} \end{bmatrix}^T$ are the states and $\mathbf{w} = F_w$ is the disturbance.

The state transition is governed by $\mathbf{A_p} = \begin{bmatrix} a_{11} & a_{12} \\ a_{21} & a_{22} \end{bmatrix}$, $\mathbf{B_p} = \begin{bmatrix} b_{11} & b_{12} & b_{13} & b_{14} \\ b_{21} & b_{22} & b_{23} & b_{24} \end{bmatrix}$ and $\mathbf{D_p} = \begin{bmatrix} d_1 \\ d_2 \end{bmatrix}$. System outputs $\mathbf{y_1} = \mathbf{y_2} = \begin{bmatrix} \mathbf{x_p} & \mathbf{u} \end{bmatrix}^T$ are governed by $\mathbf{C_{p_1}} = \mathbf{C_{p_2}} = \begin{bmatrix} \mathbf{I}_{2\times2} \\ \mathbf{0}_{4\times2} \end{bmatrix}$, $\mathbf{B_{y_1}} = \mathbf{B_{y_2}} = \begin{bmatrix} \mathbf{0}_{2\times4} \\ \mathbf{I}_{4\times4} \end{bmatrix}$ and $\mathbf{D_y} = \mathbf{0}_{6\times1}$. The measurements comprise full-state feedback with $\mathbf{M_p} = \mathbf{I}_{2\times2}$ and $\mathbf{D_z} = \mathbf{0}_{2\times1}$.

Considering the sim2real gap in terms of uncertainties in frictional coefficients of the 4 road-tire interconnects $\rho = \left< \mu_{FL}, \mu_{FR}, \mu_{RL}, \mu_{RR} \right>$, where parameters $\rho_i$ can be time-varying or constant but uncertain, this work adopts polytopic modeling method for uncertainty treatment. To this end, a polytopic LPV system can be established (refer Eq. \ref{eq_3_2}), where the state-space matrices depend affinely on the uncertain parameters.
\begin{equation}
\begin{matrix}
\mathbf{A_p\left ( \rho \right )} = \mathbf{A_0} + \mu_{FL}\mathbf{A_1} + \mu_{FR}\mathbf{A_2} + \mu_{RL}\mathbf{A_3} + \mu_{RR}\mathbf{A_4} \\
\mathbf{B_p\left ( \rho \right )} = \mathbf{B_0} + \mu_{FL}\mathbf{B_1} + \mu_{FR}\mathbf{B_2} + \mu_{RL}\mathbf{B_3} + \mu_{RR}\mathbf{B_4}
\end{matrix}
\label{eq_3_2}
\end{equation}

\textbf{\textit{Definition 1:}} A polytope of $\mathbf{S_1, S_2, ..., S_k}$ ``vertex'' systems could be represented as the convex hull of a fixed number of matrices $\mathbf{S_i}$ with the same dimension \cite{APKARIAN19951251}, i.e.,
\begin{equation}
\mathbf{Co}\left\{ \mathbf{S_i}; i \in 1, ..., k \right\} := \left\{ \sum_{i=1}^{k} \lambda_i\mathbf{S_i}: \lambda_i \geqslant 0, \sum_{i=1}^{k} \lambda_i = 1 \right\}
\label{eq_3_3}
\end{equation}

The uncertain frictional coefficients are pragmatically assumed to be bounded with $ \mu_j \in \left [ 0.1, 1.0 \right ]$ and range over a fixed polytope (refer Eq. \ref{eq_3_3}) with $k = 1, 2, ..., 16$ vertices corresponding to the $2^4$ combinations of extremal parameter values, thereby encompassing all possible values of the uncertain parameters. The resulting polytopic state-space model can be written by lumping Eq. \ref{eq_3_1} to obtain Eq. \ref{eq_3_4}, where $\mathbf{\widetilde{A}} = \mathbf{A_p}$, $\mathbf{\widetilde{B}} = \begin{bmatrix} \mathbf{D_p} & \mathbf{B_p} \end{bmatrix}$, $\mathbf{\widetilde{C}} = \begin{bmatrix} \mathbf{C_{p_1}} \\ \mathbf{C_{p_2}} \end{bmatrix}$ and $\mathbf{\widetilde{D}} = \begin{bmatrix} \mathbf{D_y} & \mathbf{B_{y_1}} \\ \mathbf{0}_{6\times1} & \mathbf{B_{y_2}} \end{bmatrix}$.
\begin{equation}
\underset{S(\rho)}{\underbrace{
\begin{bmatrix}
\mathbf{\widetilde{A}(\rho)} & \mathbf{\widetilde{B}(\rho)} \\
\mathbf{\widetilde{C}(\rho)} & \mathbf{\widetilde{D}(\rho)} \\
\end{bmatrix}}}
\in
\mathbf{Co}\left\{
\underset{S_1}{\underbrace{
\begin{bmatrix}
\mathbf{\widetilde{A}_1} & \mathbf{\widetilde{B}_1} \\
\mathbf{\widetilde{C}_1} & \mathbf{\widetilde{D}_1} \\
\end{bmatrix}}}, ...,
\underset{S_k}{\underbrace{
\begin{bmatrix}
\mathbf{\widetilde{A}_k} & \mathbf{\widetilde{B}_k} \\
\mathbf{\widetilde{C}_k} & \mathbf{\widetilde{D}_k} \\
\end{bmatrix}}}
\right\}
\label{eq_3_4}
\end{equation}

\textbf{\textit{Definition 2:}} A convex subset $\mathbf{R}$ of a complex plane is called an $n^{th}$ order LMI region if there exist a real symmetric matrix $\mathbf{L} \in \mathbb{R}^{n \times n}$ and a real matrix $\mathbf{M} \in \mathbb{R}^{n \times n}$ \cite{li2002robust}, which satisfy the LMI in $z$ and $\overline{z}$ as depicted in Eq. \ref{eq_3_5}.
\begin{equation}
\mathbf{R} = \left\{ z \in \mathbb{C}: \mathbf{L} + \mathbf{M}z + \mathbf{M}^T\overline{z} < 0 \right\}
\label{eq_3_5}
\end{equation}

\textbf{\textit{Lemma 1:}} A real matrix $\mathbf{A} \in \mathbb{R}^{n \times n}$ is $D$-stable, that is, all of its eigenvalues are in the LMI region $\mathbf{R}$ defined by Eq. \ref{eq_3_5}, if and only if there exists a positive-definite symmetric matrix $\mathbf{X} \in \mathbb{R}^{n \times n}$, which satisfies the LMI presented in Eq. \ref{eq_3_6}.
\begin{equation}
\mathbf{L} \otimes \mathbf{X} + \mathbf{M} \otimes (\mathbf{A}\mathbf{X}) + \mathbf{M}^T \otimes (\mathbf{A}\mathbf{X})^T < 0
\label{eq_3_6}
\end{equation}

We intend to place poles of the closed-loop system $S_{cl}$ in the LMI region governed by the intersection of left half-plane with $\Re(z) < \alpha$ (i.e., $\alpha$-stability), where $\alpha = -0.1$, and a conic sector centered at the origin having an inner angle of $\phi = 3\pi/4$ so as to guarantee some minimum decay rate and closed-loop damping. The LMI in Eq. \ref{eq_3_7a} represents the $\alpha$-stability region, while the one in Eq. \ref{eq_3_7b} represents the conic sector centered at the origin with an inner angle of $\phi = 2 \theta$.

\begin{subequations}
\begin{equation}
2 \alpha \mathbf{X} + \mathbf{A}\mathbf{X} + (\mathbf{A}\mathbf{X})^T < 0
\label{eq_3_7a}
\end{equation}
\begin{equation}
\begin{matrix}
\begin{bmatrix}
a \left (\mathbf{A}\mathbf{X} + (\mathbf{A}\mathbf{X})^T \right ) &  -b \left (\mathbf{A}\mathbf{X} - (\mathbf{A}\mathbf{X})^T \right )\\
b \left (\mathbf{A}\mathbf{X} - (\mathbf{A}\mathbf{X})^T \right ) &  a \left (\mathbf{A}\mathbf{X} + (\mathbf{A}\mathbf{X})^T \right )\\
\end{bmatrix}
\\
\textrm{where}, 0 < \theta < \frac{\pi}{2}, \cos(\theta) = \frac{-b}{\sqrt{a^2+b^2}}, \sin(\theta) = \frac{a}{\sqrt{a^2+b^2}} 
\end{matrix}
\label{eq_3_7b}
\end{equation}
\end{subequations}

\textbf{\textit{Definition 3:}} Energy-to-energy gain (or induced $L_2$ gain) is a performance measure of system's response $\mathbf{y}$ to disturbances $\mathbf{w}$, quantifying the amplification of the energy of input disturbances to the energy of the system outputs, and forms the supremum or $H_\infty$ norm of the system with transfer matrix $\mathbf{G}$ (i.e., largest singular value of the transfer matrix) as depicted in Eq. \ref{eq_3_8}.
\begin{equation}
\Gamma_{ee} =  \left\| \mathbf{G} \right\|_{H_\infty} = \underset{\omega\in\mathbb{R}}{\sup}\: \sigma_{\max} \left ( \mathbf{G}(j\omega) \right ) = \underset{\mathbf{w}\neq 0}{\max}\: \frac{\left\| \mathbf{y} \right\|_{L_2}}{\left\| \mathbf{w} \right\|_{L_2}}
\label{eq_3_8}
\end{equation}

\textbf{\textit{Lemma 2:}} Considering the polytopic system described by Eq. \ref{eq_3_4}, the following statements are equivalent.
\begin{enumerate}
\item The system is stable with a quadratic $H_\infty$ performance index $\gamma_1$.
\item There exists a matrix $\mathbf{P} > 0$ such that the following LMI has a feasible solution:
\begin{equation}
\begin{bmatrix}
\mathbf{P\widetilde{A}}(\rho) + \mathbf{\widetilde{A}}^T(\rho)\mathbf{P} & \mathbf{P\widetilde{B}}(\rho) & \mathbf{\widetilde{C}}^T(\rho) \\
\ast & -\gamma_1\mathbf{I} & \mathbf{\widetilde{D}}^T(\rho) \\
\ast & \ast & -\gamma_1\mathbf{I} \\
\end{bmatrix}
< 0
\label{eq_3_9}
\end{equation}
\end{enumerate}

\textbf{\textit{Definition 4:}} Energy-to-peak gain is a measure of system's response $\mathbf{y}$ to disturbances $\mathbf{w}$, quantifying the energy amplification of disturbances to their peak values as they propagate through the system to the outputs (i.e., energy of the impulse response), and forms the generalized $H_2$ norm of the system with transfer matrix $\mathbf{G}$ as depicted in Eq. \ref{eq_3_11}.
\begin{equation}
\begin{matrix}
\left\| \mathbf{G} \right\|_{H_2}^2 = \frac{1}{2\pi} \int_{-\infty}^{\infty} \mathrm{trace} \left [ \mathbf{G^*}(j\omega) \mathbf{G}(j\omega) \right ] d\omega \\
\Gamma_{ep} = \left\| \mathbf{G} \right\|_{H_2} = \underset{\mathbf{w}\neq 0}{\max}\: \frac{\left\| \mathbf{y} \right\|_{L_\infty}}{\left\| \mathbf{w} \right\|_{L_2}}
\end{matrix}
\label{eq_3_11}
\end{equation}

\textbf{\textit{Lemma 3:}} Following statements are equivalent for the uncertain system described by the polytopic model in Eq. \ref{eq_3_4}.
\begin{enumerate}
\item The system is stable with quadratic $H_2$ performance $\gamma_2$.
\item There exist $\mathbf{P} > 0$ and $\mathbf{Q} > 0$ such that:
\begin{equation}
\begin{matrix}
\mathrm{trace}\left [ \mathbf{\widetilde{C}}(\rho) \mathbf{P} \mathbf{\widetilde{C}}^T(\rho) \right ] < \gamma_2^2 \\
\mathbf{\widetilde{A}}(\rho)\mathbf{P} + \mathbf{P}\mathbf{\widetilde{A}}^T(\rho) + \mathbf{\widetilde{B}}(\rho) \mathbf{\widetilde{B}}^T(\rho) < 0 \\
\mathrm{trace}\left [ \mathbf{\widetilde{B}}^T(\rho) \mathbf{Q} \mathbf{\widetilde{B}}(\rho) \right ] < \gamma_2^2 \\
\mathbf{\widetilde{A}}^T(\rho)\mathbf{Q} + \mathbf{Q}\mathbf{\widetilde{A}}(\rho) + \mathbf{\widetilde{C}}^T(\rho) \mathbf{\widetilde{C}}(\rho) < 0
\end{matrix}
\label{eq_3_12}
\end{equation}
\end{enumerate}

\textbf{\textit{Lemma 4:}} Given matrices $\mathbf{L}$, $\mathbf{B}$ and $\mathbf{Q}$, the inequality $\mathbf{BKL}+\mathbf{L}^T\mathbf{K}^T\mathbf{B}^T+\mathbf{Q}<0$ has a solution for $\mathbf{K}$ if and only if the conditions presented in Eq. \ref{eq_3_14} are satisfied.
\begin{equation}
\begin{matrix}
\mathbf{B}^\perp \mathbf{Q} \mathbf{B}^{\perp^T} < 0 \\
\mathbf{L}^{T^\perp} \mathbf{Q} \mathbf{L}^{T^{\perp^T}} < 0
\end{matrix}
\label{eq_3_14}
\end{equation}

It can be shown that applying Eq. \ref{eq_3_14} to Eq. \ref{eq_3_9}
, we can arrive at Eq. \ref{eq_3_15a}-\ref{eq_3_15d} and much in the same way, by applying Eq. \ref{eq_3_14} to Eq. \ref{eq_3_12}
, we can obtain Eq. \ref{eq_3_16a}-\ref{eq_3_16d}.

\textbf{\textit{Theorem 1:}} For a given plant of order $n_p$, there exists an $H_\infty$ controller of order $n_c \leqslant n_p$ to stabilize the closed-loop system and guarantees $\Gamma_{ee} < \gamma_1$ if and only if the conditions presented in Eq. \ref{eq_3_15a}-\ref{eq_3_15d} are satisfied.
\scriptsize
\begin{subequations}

\begin{equation}
\begin{bmatrix}
\mathbf{B_p} \\ \mathbf{B_{y_1}}
\end{bmatrix}^\perp
\begin{bmatrix}
\mathbf{A_p X}+\mathbf{X}\mathbf{A_p}^T+\mathbf{D_p}\mathbf{D_p}^T & \mathbf{X}\mathbf{C_{p_1}}^T+\mathbf{D_p}\mathbf{D_y}^T \\
\ast & \mathbf{D_y}\mathbf{D_y}^T - \gamma_1^2\mathbf{I}\\
\end{bmatrix} 
\begin{bmatrix}
\mathbf{B_p} \\ \mathbf{B_{y_1}}
\end{bmatrix}^{\perp^T}
\label{eq_3_15a}
\end{equation}

\begin{equation}
\begin{bmatrix}
\mathbf{M_p}^T \\ \mathbf{D_z}^T
\end{bmatrix}^\perp
\begin{bmatrix}
\mathbf{Y A_p}+\mathbf{A_p}^T\mathbf{Y}+\mathbf{C_{p_1}}^T\mathbf{C_{p_1}} & \mathbf{Y}\mathbf{D_p}+\mathbf{C_{p_1}}^T\mathbf{D_y} \\
\ast & \mathbf{D_y}^T\mathbf{D_y} - \gamma_1^2\mathbf{I}\\
\end{bmatrix} 
\begin{bmatrix}
\mathbf{M_p}^T \\ \mathbf{D_z}^T
\end{bmatrix}^{\perp^T}
\label{eq_3_15b}
\end{equation}

\begin{equation}
\begin{bmatrix}
\mathbf{X} & \gamma_1\mathbf{I} \\
\gamma_1\mathbf{I} & \mathbf{Y} \\
\end{bmatrix} \geqslant 0
\label{eq_3_15c}
\end{equation}

\begin{equation}
\mathrm{rank}
\begin{bmatrix}
\mathbf{X} & \gamma_1\mathbf{I} \\
\gamma_1\mathbf{I} & \mathbf{Y} \\
\end{bmatrix} \leqslant n_p + n_c
\label{eq_3_15d}
\end{equation}

\end{subequations}
\normalsize



\textbf{\textit{Remark 1:}} For the full-state feedback controller, i.e., $n_c = 0$ and $\mathbf{M_p} = \mathbf{I}$, constraints imposed by Eq. \ref{eq_3_15b}-\ref{eq_3_15d} are satisfied, and the resultant overall problem becomes convex.

\textbf{\textit{Theorem 2:}} For a given plant of order $n_p$, there exists an $H_2$ controller of order $n_c \leqslant n_p$ to stabilize the closed-loop system and guarantees $\Gamma_{ep} < \gamma_2$ if and only if the conditions presented in Eq. \ref{eq_3_16a}-\ref{eq_3_16d} are satisfied.
\scriptsize
\begin{subequations}

\begin{equation}
\begin{cases}
\mathbf{B_p}^\perp \left ( \mathbf{A_p X}+\mathbf{X}\mathbf{A_p}^T+\mathbf{D_p}\mathbf{D_p}^T \right ) \mathbf{B_p}^{\perp^T} < 0 \\
\mathbf{C_{p_2}}\mathbf{X}\mathbf{C_{p_2}}^T < \gamma_2^2 \mathbf{I}
\end{cases}
\label{eq_3_16a}
\end{equation}

\begin{equation}
\begin{bmatrix}
\mathbf{M_p}^T \\ \mathbf{D_z}^T
\end{bmatrix}^\perp
\begin{bmatrix}
\mathbf{Y A_p}+\mathbf{A_p}^T\mathbf{Y} & \mathbf{Y}\mathbf{D_p} \\
\ast & -\mathbf{I}\\
\end{bmatrix} 
\begin{bmatrix}
\mathbf{M_p}^T \\ \mathbf{D_z}^T
\end{bmatrix}^{\perp^T}
\label{eq_3_16b}
\end{equation}

\begin{equation}
\begin{bmatrix}
\mathbf{X} & \mathbf{I} \\
\mathbf{I} & \mathbf{Y} \\
\end{bmatrix} \geqslant 0
\label{eq_3_16c}
\end{equation}

\begin{equation}
\mathrm{rank}
\begin{bmatrix}
\mathbf{X} & \mathbf{I} \\
\mathbf{I} & \mathbf{Y} \\
\end{bmatrix} \leqslant n_p + n_c
\label{eq_3_16d}
\end{equation}

\end{subequations}
\normalsize



\textbf{\textit{Remark 2:}} For the full-state feedback controller, i.e., $n_c = 0$ and $\mathbf{M_p} = \mathbf{I}$, constraints imposed by Eq. \ref{eq_3_16b}-\ref{eq_3_16d} are satisfied, and the resultant overall problem becomes convex.

From the solution $\left \{ \mathbf{X}, \mathbf{Y} \right \}$ to the convex optimization problem of minimizing $\mathrm{trace}\left ( \mathbf{X} + \mathbf{Y} \right )$ subject to Eq. \ref{eq_3_15a} and Eq. \ref{eq_3_16a}, we can build the Lyapunov matrix and obtain the optimal state-feedback controller $\mathbf{K}$ whose purpose is to minimize the influence of disturbance $\mathbf{w}$ on the uncertain system response $\mathbf{y} = \begin{bmatrix} \mathbf{y_1} & \mathbf{y_2} \end{bmatrix}^T$ resulting in a closed-loop system $S_{cl}$ with $\mathbf{A_{cl}} = \left ( \mathbf{A_p} + \mathbf{B_p}\mathbf{K} \right )$. In other words, we obtain an optimal feedback matrix $\mathbf{K}$ that achieves the following.
\begin{itemize}
\item Places the poles of $S_{cl}$ within the intersection of LMI regions $\mathbf{R}$ specified by Eq. \ref{eq_3_7a} and Eq. \ref{eq_3_7b}.
\item Bounds $\Gamma_{ee}$ of $S_{cl}$ from $\mathbf{w}$ to $\mathbf{y_1}$ below $\gamma_1 > 0$.
\item Bounds $\Gamma_{ep}$ of $S_{cl}$ from $\mathbf{w}$ to $\mathbf{y_2}$ below $\gamma_2 > 0$.
\item Minimizes the $H_2$-$H_\infty$ tradeoff criterion of the form $\phi \Gamma_{ee}^2 + \varphi \Gamma_{ep}^2$ with $\phi$ and $\varphi$ being weights on the respective performance measures.
\end{itemize}


\textbf{\textit{Remark 3:}} With the specific choice of output signals $\mathbf{y} = \begin{bmatrix} \mathbf{y_1} & \mathbf{y_2} \end{bmatrix}^T$, the designed controller minimizes the deviation of state $\mathbf{x_p}$ from zero (for stabilization) or reference $\mathbf{x_r}$ (for tracking) along with the control inputs $\mathbf{u}$. From a sim2real transfer perspective, the devised framework guarantees stabilizing/tracking performance in the presence of parameter uncertainties and disturbances while also minimizing the control effort.


\section{Results and Discussion}
\label{Section: Results and Discussion}

We first established the benefit of adopting an unconventional vehicle architecture by comparing it against a conventional one with the same parameters (refer Table \ref{tab2}). The conventional Ackermann-steered architecture yielded a Pareto-optimal $H_\infty$ performance of 2.61e-1 and $H_2$ performance of 6.45e-1, while the unconventional independent 4WD4WS architecture was able to guarantee $H_\infty$ performance of 1.98e-01 and $H_2$ performance of 5.56e-01. In other words, the conventional system was poorly robust as compared to the proposed one and was, therefore, capable of handling relatively narrower sim2real gaps. Consequently, further results are reported only for the independent 4WD4WS vehicle architecture.

\begin{table}[b]
\centering
\caption{Benchmarking analysis of the proposed sim2real framework}
\label{tab3}
\resizebox{\columnwidth}{!}{%
\begin{tabular}{l|ll|ll|ll}
\hline
\multirow{2}{*}{Error ($\varepsilon$) $\searrow$} & \multicolumn{2}{c|}{Open-Loop}            & \multicolumn{2}{c|}{Non-Robust}          & \multicolumn{2}{c}{Robust}               \\ \cline{2-7} 
                                    & \multicolumn{1}{c|}{Sim}       & \multicolumn{1}{c|}{Real}     & \multicolumn{1}{c|}{Sim}      & \multicolumn{1}{c|}{Real}     & \multicolumn{1}{c|}{Sim}      & \multicolumn{1}{c}{Real}     \\ \hline
Straight                            & \multicolumn{1}{c|}{3.81e-2}  & 5.03e-2 & \multicolumn{1}{c|}{9.73e-2} & 6.04e-2 & \multicolumn{1}{c|}{1.83e-2} & 1.82e-2 \\
Lane-Change                         & \multicolumn{1}{c|}{7.59e-2} & 6.59e-2 & \multicolumn{1}{c|}{5.78e-2} & 3.38e-2 & \multicolumn{1}{c|}{1.20e-2} & 1.18e-2 \\
Skidpad                             & \multicolumn{1}{c|}{1.09e-1} & 1.01e-1 & \multicolumn{1}{c|}{7.91e-2} & 8.36e-2 & \multicolumn{1}{c|}{2.14e-2} & 2.06e-2 \\
Fishhook                            & \multicolumn{1}{c|}{5.52e-2}  & 6.79e-2 & \multicolumn{1}{c|}{4.23e-2} & 7.75e-2 & \multicolumn{1}{c|}{1.50e-2} & 1.82e-2 \\
Slalom                              & \multicolumn{1}{c|}{1.82e-1}  & 1.91e-1 & \multicolumn{1}{c|}{2.87e-1} & 2.52e-1 & \multicolumn{1}{c|}{6.14e-2} & 6.15e-2 \\
Figure-8                            & \multicolumn{1}{c|}{1.43e-1}  & 4.15e-1 & \multicolumn{1}{c|}{6.85e-2} & 5.47e-1 & \multicolumn{1}{c|}{4.67e-2} & 1.07e-1 \\ \hline
\end{tabular}%
}
\end{table}


Next, we benchmarked the performance of the proposed robust sim2real control framework against (i) replayed reference control commands in open-loop and (ii) another non-robust state-feedback controller devised with a pole-placement approach based on Eq. \ref{eq_3_7a} to ensure $\Re(z) < -2.0$. Particularly, we analyzed the root-mean-squared error (RMSE), $\epsilon = \begin{bmatrix} \epsilon_x \; \epsilon_y \; \epsilon_\psi \\ \end{bmatrix}^T$, for the Cartesian pose of the vehicle throughout a given trajectory. The single error metric reported in Table \ref{tab3} is the $\ell^2$-norm of the resulting errors in each of the pose coordinates, i.e., $\varepsilon = \left\|\epsilon\right\|_2$.

\begin{figure*}[t]
	\centering
	\includegraphics[width=\linewidth]{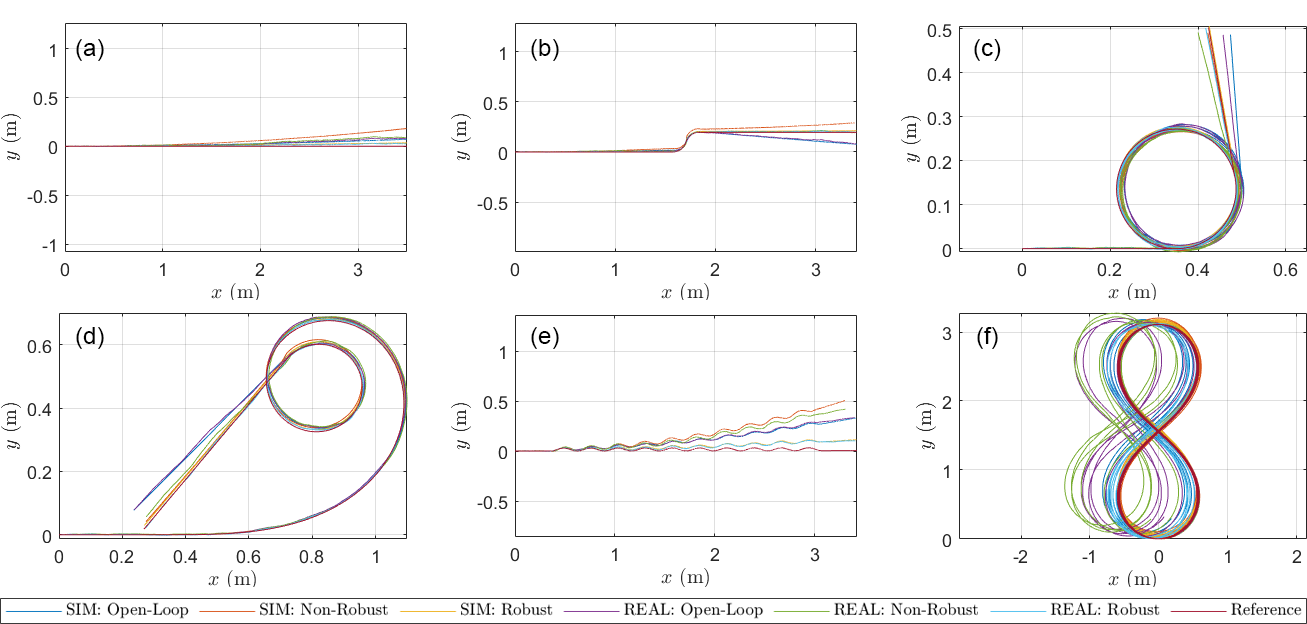}
	\caption{Experimentation and benchmarking results from simulation and real-world for (a) stabilizing as well as (b)-(f) steady-state tracking of standard maneuvers, viz. (b) lane-change, (c) skidpad, (d) fishhook, (e) slalom tests, and (f) figure-8, under deliberately injected exogenous disturbances and uncertainties.}
	\label{fig7}
\end{figure*}

The design of experiments followed a common approach for simulation and real-world settings. The stabilizing controller was tested for a straight-line maneuver (refer Fig. \hyperref[fig5]{\ref*{fig7}(a)}). The steady-state tracking control was validated for standard benchmark maneuvers, viz. lane-change (refer Fig. \hyperref[fig5]{\ref*{fig7}(b)}), skidpad (refer Fig. \hyperref[fig5]{\ref*{fig7}(c)}), fishhook (refer Fig. \hyperref[fig5]{\ref*{fig7}(d)}), slalom (refer Fig. \hyperref[fig5]{\ref*{fig7}(e)}) and figure-8 (refer Fig. \hyperref[fig5]{\ref*{fig7}(f)}), wherein the reference generation was achieved by simulating the open-loop system with standard test signals, viz. impulse, step, ramp and sine.

In simulation experiments, the friction for individual road-tire interconnects was exaggerated as phase-shifted sinusoids with $\mu_i =  0.35 \sin(2\pi f \theta-\phi) + 0.55 \in \left [ 0.2,0.9 \right ]$, where $f$=12 is the frequency of variation, $\theta$=$[0,2\pi]$ is the range of variation and $\phi$ is the phase lag. The resulting sinusoids were injected with synthetic noise defined as $\mu_i -0.05 + 0.1\chi$, where $\chi \sim U(0,1)$ is a uniform random variable. Here, $\mu_{FR}$, $\mu_{RL}$ and $\mu_{RR}$ respectively lagged $\mu_{FL}$ by $\phi_{FR}=\pi/2$, $\phi_{RL}=\pi$ and $\phi_{RR}=3\pi/2$ rad, such that all four wheels never experienced the same $\mu$. Additionally, noisy wind-gust disturbance $F_w$ of $0.25$ N amplitude was simulated as a step input from $t$=1 second onwards, with synthetic noise of the form $F_w + 0.1\chi*\mathrm{max}(F_w)$, where $\chi \sim U(0,1)$ is a uniform random variable.


In real-world experiments, the vehicle state feedback was obtained using OptiTrack motion capture system. Although a nominal value of friction coefficient was estimated to be 0.4 from the motion capture data (refer Table \ref{tab2}), the test surface (5.0$\times$3.5 m) was deliberately left unclean with dust/sand particles, irregular scratches and tape residues from previous experiments. Additionally, soap-water solutions of varying concentrations were spilled in uneven quantities to aggravate the uncertainty in terms of road friction. Finally, in the figure-8 maneuver (depicted as a freeze-frame sequence in Fig. \ref{fig1}), the vehicle was actively poked with a pole at $t$=52, $t$=68, $t$=81 and $t$=109 seconds, which acted as an impulse disturbance. The exact values of real-world friction and disturbance were time-varying and very difficult to measure and were therefore treated to be a part of the ``unknown'' sim2real gap.

Quantitatively, the proposed robust sim2real framework has the least error across all maneuvers for all experiments. Additionally, the difference between ``Sim'' and ``Real'' error metrics practically reflects the sim2real gap, wherein the proposed framework clearly outperforms its counterparts.

Qualitatively, all the systems track the first lane-change transient relatively closely, but the open-loop and non-robust systems digress significantly after the second one. For skidpad and fishhook maneuvers, all the system performances overlap before and within the loops, albeit with slight variations, after which we can see a drastic variation, as different systems exit at different angles. For straight and slalom maneuvers, the open-loop system deceptively seems to outperform the non-robust closed-loop system, but note that it is not actively correcting any errors. Finally, figure-8 maneuver was the only one where the real-world system was actively disturbed and, consequently, the simulated and real-world performance have a wider gap in this case. Nevertheless, the proposed framework still outperforms the open-loop and non-robust systems.


\section{Conclusion}
\label{Section: Conclusion}

In this work, we first introduced the complete mechatronic design architecture of Nigel and also presented the detailed dynamics modeling of this independent 4WD4WS vehicle. We also formulated a linear parameter-varying (polytopic) model of the system for synthesizing a robust optimal controller, which seeks to minimize the $H_2$-$H_\infty$ tradeoff criteria with $D$-stability guarantee. We demonstrated and analyzed robust stabilizing as well as steady-state tracking control of Nigel with extensive experimentation and benchmarking in simulation as well as real world. In addition to sim2real transfer, we also validated the controller to handle exaggerated disturbance and uncertainties in real-world conditions.

Future work will delve into formulating and validating a robust tracking problem using mixed sensitivity loop-shaping. Other avenues include full-scale deployment, fault-tolerant control, and integration of the proposed framework with an end-to-end autonomy stack.


\bibliographystyle{IEEEtran}
\bibliography{IEEEabrv,References}

\begin{thebibliography}{10}
\providecommand{\url}[1]{#1}
\csname url@rmstyle\endcsname
\providecommand{\newblock}{\relax}
\providecommand{\bibinfo}[2]{#2}
\providecommand\BIBentrySTDinterwordspacing{\spaceskip=0pt\relax}
\providecommand\BIBentryALTinterwordstretchfactor{4}
\providecommand\BIBentryALTinterwordspacing{\spaceskip=\fontdimen2\font plus
\BIBentryALTinterwordstretchfactor\fontdimen3\font minus \fontdimen4\font\relax}
\providecommand\BIBforeignlanguage[2]{{%
\expandafter\ifx\csname l@#1\endcsname\relax
\typeout{** WARNING: IEEEtran.bst: No hyphenation pattern has been}%
\typeout{** loaded for the language `#1'. Using the pattern for}%
\typeout{** the default language instead.}%
\else
\language=\csname l@#1\endcsname
\fi
#2}}

\bibitem{deSilva2004}
\BIBentryALTinterwordspacing
C.~de~Silva, \emph{{Mechatronics: An Integrated Approach}}.\hskip 1em plus 0.5em minus 0.4em\relax Taylor \& Francis, 2004. [Online]. Available: \url{https://books.google.com/books?id=CjB2ygeR95cC}
\BIBentrySTDinterwordspacing

\bibitem{CS4AV}
C.~V. Samak, T.~V. Samak, and S.~Kandhasamy, ``{Control Strategies for Autonomous Vehicles},'' in \emph{Autonomous Driving and Advanced Driver-Assistance Systems (ADAS)}.\hskip 1em plus 0.5em minus 0.4em\relax CRC Press, 2021, pp. 37--86.

\bibitem{MIT-Racecar2017}
\BIBentryALTinterwordspacing
S.~Karaman~et al., ``{Project-based, collaborative, algorithmic robotics for high school students: Programming self-driving race cars at MIT},'' in \emph{2017 IEEE Integrated STEM Education Conference (ISEC)}, 2017, pp. 195--203. [Online]. Available: \url{https://mit-racecar.github.io}
\BIBentrySTDinterwordspacing

\bibitem{AutoRally2021}
\BIBentryALTinterwordspacing
B.~Goldfain, P.~Drews, C.~You, M.~Barulic, O.~Velev, P.~Tsiotras, and J.~M. Rehg, ``{AutoRally: An Open Platform for Aggressive Autonomous Driving},'' \emph{IEEE Control Systems Magazine}, vol.~39, no.~1, pp. 26--55, 2019. [Online]. Available: \url{https://arxiv.org/abs/1806.00678}
\BIBentrySTDinterwordspacing

\bibitem{F1TENTH2019}
\BIBentryALTinterwordspacing
M.~O'Kelly, V.~Sukhil, H.~Abbas, J.~Harkins, C.~Kao, Y.~V. Pant, R.~Mangharam, D.~Agarwal, M.~Behl, P.~Burgio, and M.~Bertogna. (2019) {F1/10: An Open-Source Autonomous Cyber-Physical Platform}. [Online]. Available: \url{https://arxiv.org/abs/1901.08567}
\BIBentrySTDinterwordspacing

\bibitem{DSV2017}
\BIBentryALTinterwordspacing
T.~K. et~al., ``{Design and Development of the Delft Scaled Vehicle: A Platform for Autonomous Driving Tests},'' Delft University of Technology, Delft, Netherlands, Bachelor's Thesis, 2017. [Online]. Available: \url{https://www.erwinrietveld.com/assets/docs/BEP11\_DSCS\_Paper\_Final.pdf}
\BIBentrySTDinterwordspacing

\bibitem{MuSHR2019}
\BIBentryALTinterwordspacing
S.~S. Srinivasa, P.~Lancaster, J.~Michalove, M.~Schmittle, C.~Summers, M.~Rockett, J.~R. Smith, S.~Choudhury, C.~Mavrogiannis, and F.~Sadeghi. (2019) {MuSHR: A Low-Cost, Open-Source Robotic Racecar for Education and Research}. [Online]. Available: \url{https://arxiv.org/abs/1908.08031}
\BIBentrySTDinterwordspacing

\bibitem{BARC2021}
\BIBentryALTinterwordspacing
J.~Pappas, C.~H. Yuan, C.~S. Lu, N.~Nassar, A.~Miller, S.~van Leeuwen, and F.~Borrelli, ``{Berkeley Autonomous Race Car (BARC)},'' 2021. [Online]. Available: \url{https://sites.google.com/site/berkeleybarcproject}
\BIBentrySTDinterwordspacing

\bibitem{ORCA2021}
\BIBentryALTinterwordspacing
{Automatic Control Laboratory, ETH Zürich}, ``{ORCA (Optimal RC Racing) Project},'' 2021. [Online]. Available: \url{https://control.ee.ethz.ch/research/team-projects/autonomous-rc-car-racing.html}
\BIBentrySTDinterwordspacing

\bibitem{HyphaROS-Racecar2021}
\BIBentryALTinterwordspacing
{HyphaROS Workshop}, ``{HyphaROS Racecar},'' 2021. [Online]. Available: \url{https://github.com/Hypha-ROS/hypharos\_racecar}
\BIBentrySTDinterwordspacing

\bibitem{DonkeyCar2021}
\BIBentryALTinterwordspacing
{Donkey Community}, ``{An Open-Source DIY Self-Driving Platform for Small-Scale Cars},'' 2021. [Online]. Available: \url{https://www.donkeycar.com}
\BIBentrySTDinterwordspacing

\bibitem{QCar2021}
\BIBentryALTinterwordspacing
{Quanser Consulting Inc.}, ``{QCar - A Sensor-Rich Autonomous Vehicle},'' 2021. [Online]. Available: \url{https://www.quanser.com/products/qcar}
\BIBentrySTDinterwordspacing

\bibitem{DeepRacer2021}
\BIBentryALTinterwordspacing
{Amazon Web Services}, ``{AWS DeepRacer},'' 2021. [Online]. Available: \url{https://aws.amazon.com/deepracer}
\BIBentrySTDinterwordspacing

\bibitem{Duckietown2017}
\BIBentryALTinterwordspacing
L.~Paull~et al., ``{Duckietown: An Open, Inexpensive and Flexible Platform for Autonomy Education and Research},'' in \emph{2017 IEEE International Conference on Robotics and Automation (ICRA)}, 2017, pp. 1497--1504. [Online]. Available: \url{http://michalcap.net/wp-content/papercite-data/pdf/paull\_2017.pdf}
\BIBentrySTDinterwordspacing

\bibitem{Turtlebot2021}
\BIBentryALTinterwordspacing
{Robotis Inc.}, ``{TurtleBot3},'' 2021. [Online]. Available: \url{https://emanual.robotis.com/docs/en/platform/turtlebot3/overview/}
\BIBentrySTDinterwordspacing

\bibitem{electronics12163511}
\BIBentryALTinterwordspacing
B.~Bae and D.-H. Lee, ``{Design of a Four-Wheel Steering Mobile Robot Platform and Adaptive Steering Control for Manual Operation},'' \emph{Electronics}, vol.~12, no.~16, 2023. [Online]. Available: \url{https://www.mdpi.com/2079-9292/12/16/3511}
\BIBentrySTDinterwordspacing

\bibitem{s22062144}
\BIBentryALTinterwordspacing
J.~Park and Y.~Park, ``{Multiple-Actuator Fault Isolation Using a Minimal L1-Norm Solution with Applications in Overactuated Electric Vehicles},'' \emph{Sensors}, vol.~22, no.~6, 2022. [Online]. Available: \url{https://www.mdpi.com/1424-8220/22/6/2144}
\BIBentrySTDinterwordspacing

\bibitem{AutoDRIVEMechatronics2023}
C.~Samak, T.~Samak, and V.~Krovi, ``{Towards Mechatronics Approach of System Design, Verification and Validation for Autonomous Vehicles},'' in \emph{2023 IEEE/ASME International Conference on Advanced Intelligent Mechatronics (AIM)}, 2023, pp. 1208--1213.

\bibitem{AutoDRIVEEcosystem2022}
\BIBentryALTinterwordspacing
T.~Samak, C.~Samak, S.~Kandhasamy, V.~Krovi, and M.~Xie, ``{AutoDRIVE: A Comprehensive, Flexible and Integrated Digital Twin Ecosystem for Autonomous Driving Research \& Education},'' \emph{Robotics}, vol.~12, no.~3, 2023. [Online]. Available: \url{https://www.mdpi.com/2218-6581/12/3/77}
\BIBentrySTDinterwordspacing

\bibitem{AutoDRIVEReport2021}
\BIBentryALTinterwordspacing
T.~V. Samak and C.~V. Samak, ``{AutoDRIVE - Technical Report},'' 2022. [Online]. Available: \url{https://arxiv.org/abs/2211.08475}
\BIBentrySTDinterwordspacing

\bibitem{AutoDRIVESimulator2021}
\BIBentryALTinterwordspacing
T.~V. Samak, C.~V. Samak, and M.~Xie, ``{AutoDRIVE Simulator: A Simulator for Scaled Autonomous Vehicle Research and Education},'' in \emph{2021 2nd International Conference on Control, Robotics and Intelligent System}, ser. CCRIS'21.\hskip 1em plus 0.5em minus 0.4em\relax New York, NY, USA: Association for Computing Machinery, 2021, p. 1–5. [Online]. Available: \url{https://doi.org/10.1145/3483845.3483846}
\BIBentrySTDinterwordspacing

\bibitem{AutoDRIVESimulatorReport2020}
\BIBentryALTinterwordspacing
T.~V. Samak and C.~V. Samak, ``{AutoDRIVE Simulator - Technical Report},'' 2022. [Online]. Available: \url{https://arxiv.org/abs/2211.07022}
\BIBentrySTDinterwordspacing

\bibitem{Zhang2021}
\BIBentryALTinterwordspacing
Z.~Zhang, C.~Yang, W.~Zhang, Y.~Xu, Y.~Peng, and M.~Chi, ``{Motion Control of a 4WS4WD Path-Following Vehicle: Dynamics-Based Steering and Driving Models},'' \emph{Shock and Vibration}, vol. 2021, 2021. [Online]. Available: \url{https://doi.org/10.1155/2021/8861159}
\BIBentrySTDinterwordspacing

\bibitem{electronics11223731}
\BIBentryALTinterwordspacing
S.~Zhu, B.~Wei, D.~Liu, H.~Chen, X.~Huang, Y.~Zheng, and W.~Wei, ``{A Dynamics Coordinated Control System for 4WD-4WS Electric Vehicles},'' \emph{Electronics}, vol.~11, no.~22, 2022. [Online]. Available: \url{https://www.mdpi.com/2079-9292/11/22/3731}
\BIBentrySTDinterwordspacing

\bibitem{10.1115/DSCC2012-MOVIC2012-8603}
\BIBentryALTinterwordspacing
J.~R. Kolodziej, ``{Adaptive Rear-Wheel Steering Control of a Four-Wheel Vehicle Over Uncertain Terrain},'' ser. Dynamic Systems and Control Conference, vol.~1, 10 2012, pp. 857--866. [Online]. Available: \url{https://doi.org/10.1115/DSCC2012-MOVIC2012-8603}
\BIBentrySTDinterwordspacing

\bibitem{SCHWARTZ2019162}
\BIBentryALTinterwordspacing
M.~Schwartz, F.~Siebenrock, and S.~Hohmann, ``{Model Predictive Control Allocation of an Over-Actuated Electric Vehicle with Single Wheel Actuators},'' \emph{IFAC-PapersOnLine}, vol.~52, no.~8, pp. 162--169, 2019, 10th IFAC Symposium on Intelligent Autonomous Vehicles IAV 2019. [Online]. Available: \url{https://www.sciencedirect.com/science/article/pii/S2405896319303969}
\BIBentrySTDinterwordspacing

\bibitem{app8061000}
\BIBentryALTinterwordspacing
Q.~Tan, P.~Dai, Z.~Zhang, and J.~Katupitiya, ``{MPC and PSO Based Control Methodology for Path Tracking of 4WS4WD Vehicles},'' \emph{Applied Sciences}, vol.~8, no.~6, 2018. [Online]. Available: \url{https://www.mdpi.com/2076-3417/8/6/1000}
\BIBentrySTDinterwordspacing

\bibitem{1081954}
\BIBentryALTinterwordspacing
J.-E. Moseberg and G.~Roppenecker, ``{Robust Cascade Control for the Horizontal Motion of a Vehicle with Single-Wheel Actuators},'' \emph{Vehicle System Dynamics}, vol.~53, no.~12, pp. 1742--1758, 2015. [Online]. Available: \url{https://doi.org/10.1080/00423114.2015.1081954}
\BIBentrySTDinterwordspacing

\bibitem{LI2024104621}
\BIBentryALTinterwordspacing
M.~Li, Y.~Jia, and T.~Lei, ``{Path Tracking of Varying-Velocity 4WS Autonomous Vehicles under Tire Force Friction Ellipse Constraints},'' \emph{Robotics and Autonomous Systems}, vol. 173, p. 104621, 2024. [Online]. Available: \url{https://www.sciencedirect.com/science/article/pii/S0921889024000046}
\BIBentrySTDinterwordspacing

\bibitem{221348}
J.~Ackermann and W.~Sienel, ``{Robust Yaw Damping of Cars with Front and Rear Wheel Steering},'' \emph{IEEE Transactions on Control Systems Technology}, vol.~1, no.~1, pp. 15--20, 1993.

\bibitem{doi:10.1080/00423114.2013.879190}
\BIBentryALTinterwordspacing
H.~Zhang, X.~Zhang, and J.~Wang, ``{Robust Gain-Scheduling Energy-to-Peak Control of Vehicle Lateral Dynamics Stabilisation},'' \emph{Vehicle System Dynamics}, vol.~52, no.~3, pp. 309--340, 2014. [Online]. Available: \url{https://doi.org/10.1080/00423114.2013.879190}
\BIBentrySTDinterwordspacing

\bibitem{8961203}
S.~Cheng~et al., ``{Robust LMI-Based H-Infinite Controller Integrating AFS and DYC of Autonomous Vehicles With Parametric Uncertainties},'' \emph{IEEE Transactions on Systems, Man, and Cybernetics: Systems}, vol.~51, no.~11, pp. 6901--6910, 2021.

\bibitem{s21237850}
\BIBentryALTinterwordspacing
J.~P. Redondo, B.~L. Boada, and V.~Díaz, ``{LMI-Based H-Infinity Controller of Vehicle Roll Stability Control Systems with Input and Output Delays},'' \emph{Sensors}, vol.~21, no.~23, 2021. [Online]. Available: \url{https://www.mdpi.com/1424-8220/21/23/7850}
\BIBentrySTDinterwordspacing

\bibitem{Babawuro_2020}
\BIBentryALTinterwordspacing
A.~Y. Babawuro, N.~M. Tahir, M.~Muhammed, and A.~U. Sambo, ``{Optimized State Feedback Control of Quarter Car Active Suspension System Based on LMI Algorithm},'' \emph{Journal of Physics: Conference Series}, vol. 1502, no.~1, p. 012019, Mar 2020. [Online]. Available: \url{https://dx.doi.org/10.1088/1742-6596/1502/1/012019}
\BIBentrySTDinterwordspacing

\bibitem{5160361}
Y.-e. Mao, Y.~Zheng, Y.~Jing, G.~M. Dimirovski, and S.~Hang, ``{An LMI Approach to Slip Ratio Control of Vehicle Antilock Braking Systems},'' in \emph{2009 American Control Conference}, 2009, pp. 3350--3354.

\bibitem{9108078}
M.~Schwartz, T.~Rudolf, and S.~Hohmann, ``{Robust Position and Velocity Tracking Control of a Four-wheel Drive and Four-wheel Steered Electric Vehicle},'' in \emph{2020 6th International Conference on Control, Automation and Robotics (ICCAR)}, 2020, pp. 415--422.

\bibitem{SAMAK2023277}
\BIBentryALTinterwordspacing
C.~Samak, T.~Samak, and V.~Krovi, ``{Towards Sim2Real Transfer of Autonomy Algorithms using AutoDRIVE Ecosystem},'' \emph{IFAC-PapersOnLine}, vol.~56, no.~3, pp. 277--282, 2023, 3rd Modeling, Estimation and Control Conference MECC 2023. [Online]. Available: \url{https://www.sciencedirect.com/science/article/pii/S2405896323023704}
\BIBentrySTDinterwordspacing

\bibitem{milliken1995race}
\BIBentryALTinterwordspacing
W.~Milliken and D.~Milliken, \emph{{Race Car Vehicle Dynamics}}, ser. Premiere Series.\hskip 1em plus 0.5em minus 0.4em\relax SAE International, 1995. [Online]. Available: \url{https://books.google.com/books?id=EOHPjgEACAAJ}
\BIBentrySTDinterwordspacing

\bibitem{rajamani2011vehicle}
\BIBentryALTinterwordspacing
R.~Rajamani, \emph{{Vehicle Dynamics and Control}}, ser. Mechanical Engineering Series.\hskip 1em plus 0.5em minus 0.4em\relax Springer US, 2011. [Online]. Available: \url{https://books.google.com/books?id=q6SJcgAACAAJ}
\BIBentrySTDinterwordspacing

\bibitem{APKARIAN19951251}
\BIBentryALTinterwordspacing
P.~Apkarian, P.~Gahinet, and G.~Becker, ``{Self-Scheduled H-Infinity Control of Linear Parameter-Varying Systems: A Design Example},'' \emph{Automatica}, vol.~31, no.~9, pp. 1251--1261, 1995. [Online]. Available: \url{https://www.sciencedirect.com/science/article/pii/000510989500038X}
\BIBentrySTDinterwordspacing

\bibitem{li2002robust}
Y.~Li, ``{Robust Control - LMI Method},'' 2002.

\end{thebibliography}


\begin{IEEEbiography}[{\includegraphics[width=1in,height=1.25in,clip,keepaspectratio]{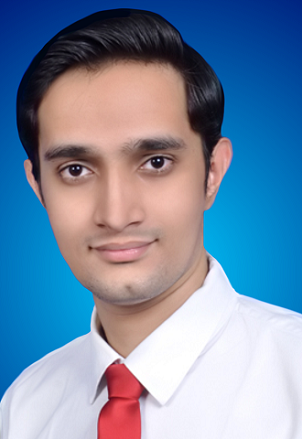}}]{Chinmay V. Samak} (Student Member, IEEE) received the B.Tech. degree in mechatronics engineering from SRM Institute of Science and Technology with a gold medal in 2021. He is currently pursuing a direct Ph.D. degree at Clemson University International Center for Automotive Research (CU-ICAR) where he is working at the ARMLab. His research interests lie at the intersection of physics-informed and data-driven methods to bridge the sim2real gap using autonomy-oriented digital twins.
\end{IEEEbiography}

\begin{IEEEbiography}[{\includegraphics[width=1in,height=1.25in,clip,keepaspectratio]{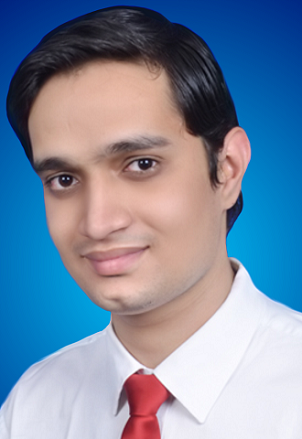}}]{Tanmay V. Samak} (Student Member, IEEE) received the B.Tech. degree in mechatronics engineering from SRM Institute of Science and Technology with a silver medal in 2021. He is currently pursuing a direct Ph.D. degree at Clemson University International Center for Automotive Research (CU-ICAR) where he is working at the ARMLab. His research interests include autonomy-oriented modeling, estimation and simulation methods aimed at bridging the real2sim gap to
develop physically and graphically accurate digital twins.
\end{IEEEbiography}

\begin{IEEEbiography}[{\includegraphics[width=1in,height=1.25in,clip,keepaspectratio]{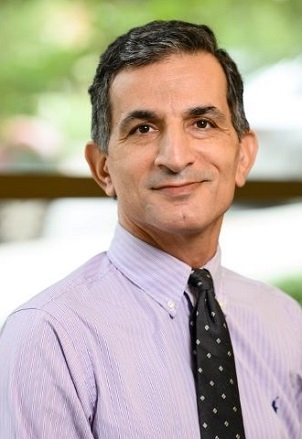}}]{Javad M. Velni} (Senior Member, IEEE) received the Ph.D. degree in mechanical engineering from University Houston in 2008. He is a professor with the Department of Mechanical Engineering at Clemson University, where he also directs the Velni Lab. His research interests include secure control of cyber–physical systems and, in particular, transportation systems, coverage control of heterogeneous multi-agent systems, and learning-based control of complex distributed systems.
\end{IEEEbiography}

\begin{IEEEbiography}[{\includegraphics[width=1in,height=1.25in,clip,keepaspectratio]{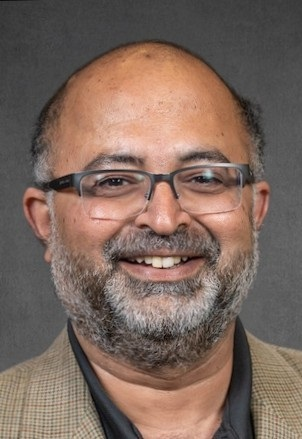}}]{Venkat N. Krovi} (Senior Member, IEEE) received the Ph.D. degree in mechanical engineering and applied mechanics from University of Pennsylvania in 1998. He is the Michelin Endowed Chair Professor of Vehicle Automation with the Departments of Automotive and Mechanical Engineering at Clemson University, where he also directs the ARMLab. The underlying theme of his research has been to take advantage of ``power of the many'' (both autonomous agents and humans) to extend the reach of human users in dull, dirty, and dangerous environments.
\end{IEEEbiography}

\vfill

\end{document}